\documentclass[runningheads]{llncs}
\usepackage{graphicx}
\usepackage{amsmath,amssymb,amsfonts} % define this before the line numbering.
\usepackage{color}
\usepackage[width=122mm,left=12mm,paperwidth=146mm,height=193mm,top=12mm,paperheight=217mm]{geometry}
\usepackage[ruled,vlined,linesnumbered]{algorithm2e}
\usepackage[percent]{overpic}
\usepackage{xcolor}
\usepackage{cite}
\usepackage{capt-of}

\begin{document}
% \linenumbers

\newcommand{\etal}{et al.}
\newcommand{\x}{\mathbf{x}}
\newcommand{\y}{\mathbf{y}}

\newcommand{\I}{\mathbf{I}}

\newcommand{\E}{\mathbf{e}}
\newcommand{\DX}{\mathbf{\Delta x}}
\newcommand{\DY}{\mathbf{\Delta y}}
\newcommand{\p}{\mathbf{p}}
\newcommand{\argmin}{\operatornamewithlimits{argmin}}
\newcommand{\mymin}{\operatornamewithlimits{min}}
\newcommand{\wxi}{\mathcal{\mathbf{w}}^{(i)}_x}
\newcommand{\wyi}{\mathcal{\mathbf{w}}^{(i)}_y}
\newcommand{\Nizero}{\mathbf{0}_{N_i }}
\newcommand{\Nione}{\mathbf{1}_{N_i }}

\newcommand*{\Scale}[2][4]{\scalebox{#1}{$#2$}}%

\newcommand\T{{\hspace{-1pt}\intercal}}
\def\eqnvspace{{\vspace{-2mm}}}
\def\figvspace{{\vspace{-5mm}}}
\newcommand{\Paragraph}[1]{\vspace{-0mm} \noindent \textbf{#1} \hspace{-1mm}}
\newcommand{\Section}[1]{\vspace{-4mm} \section{#1} \vspace{-2mm}}
\newcommand{\SubSection}[1]{\vspace{-3mm} \subsection{#1} \vspace{-2mm}}

\newcommand{\cfbox}[2]{%
    \colorlet{currentcolor}{.}%
    {\color{#1}%
    \fbox{\color{currentcolor}#2}}%
}

\pagestyle{headings}
\mainmatter
\def\ECCV16SubNumber{***}  % Insert your submission number here

\title{Temporally Robust Global Motion Compensation by Keypoint-based Congealing} % Replace with your title

\titlerunning{Temporally Robust Global Motion Compensation}

\authorrunning{S. Morteza Safdarnejad et al.}

\author{S. Morteza Safdarnejad, Yousef Atoum, Xiaoming Liu}
\institute{Michigan State University}

\maketitle

\begin{abstract}
Global motion compensation (GMC) removes the impact of camera motion and creates a video in which the background appears static over the progression of time.
Various vision problems, such as human activity recognition, background reconstruction, and multi-object tracking can benefit from GMC.
Existing GMC algorithms rely on sequentially processing consecutive frames, by estimating the transformation mapping the two frames, and obtaining a composite transformation to a global motion compensated coordinate.
Sequential GMC suffers from temporal drift of frames from the accurate global coordinate, due to either error accumulation or sporadic failures of motion estimation at a few frames. 
We propose a temporally robust global motion compensation (TRGMC) algorithm which performs accurate and stable GMC, despite complicated and long-term camera motion.
TRGMC densely connects pairs of frames, by matching local keypoints of each frame.  
A joint alignment of these frames is formulated as a novel keypoint-based congealing problem, where the transformation of each frame is updated iteratively, such that the spatial coordinates for the start and end points of matched keypoints are identical.
Experimental results demonstrate that TRGMC has superior performance in a wide range of scenarios.

\end{abstract}

\vspace{-4mm}
\Section{Introduction}
Global motion compensation (GMC) removes the impact of \textit{intentional} and \textit{unwanted} camera motion in the video, transfroming the video to have~\textit{static} background with the only motion coming from foreground objects.
%Video stabilization is a closely related problem where \textit{unwanted} camera motion, such as vibration, is removed, leaving a smooth camera motion in the output video.
As a related problem, video stabilization removes \textit{unwanted} camera motion, such as vibration, and generates a video with a \textit{smooth} camera motion.
%It is important to note that the final product of GMC is a video with~\textit{static} background~\textit{throughout the entire video}. 
%This sets a high bar on accuracy requirement for estimation of transformations to the global coordinate, despite foreground motion and appearance ambiguities.
%GMC can be re-purposed for video stabilization (VS) and mosaicing, but not vice versa - given the accuracy requirement. 
The term ``global motion compensation" is also used in video coding literature, where background motion is estimated roughly to enhance the video compression performance~\cite{he2001fast, smolic2003long}. % in some compression formats such as MPEG-4  %but very fast, 

%There are a wide range of applications for GMC.
GMC is an essential module for processing videos from {\it non-stationary} cameras,
%Its importance is apparent 
which are abundant due to emerging mobile sensors, e.g., wearable cameras, smartphones, and camera drones. %, and smartglasses.
First, the resultant \textit{motion panorama}~\cite{bartoli2002video}, as if virtually generated by a static camera, is by itself appealing for visual perception.
More importantly, many vision tasks benefit from GMC.
% and virtually creating a video which seems to be generated by a static camera.
%GMC is widely applicable both for professionally shot videos and amateur videos. 
%Considering the boom of amateur video capture devices such as smart phones, wearable cameras, camera drones, and smart glasses, GMC is found to be more significant for many applications.
For instance, dense trajectories~\cite{dense} are shown to be superior %more informative for motion analysis 
when camera motion is compensated%, either as a pre-processing step, or internally
~\cite{improveddense}.
Otherwise, camera motion interferes with human motion, rendering the analysis problem very challenging.
GMC allows reconstruction of a ``stitched" background~\cite{monari2011real}, and subsequently segmentation of foreground~\cite{sun2006better, wan2008new}. 
This helps multi-object tracking by mitigating the unconstrained problem of tracking multiple in-the-wild objects, to tracking objects with a static background~\cite{solera2015learning}.

\begin{figure} [t]
\centering
\includegraphics[trim=8cm 10cm 17cm 15cm, clip,width=.9\columnwidth]{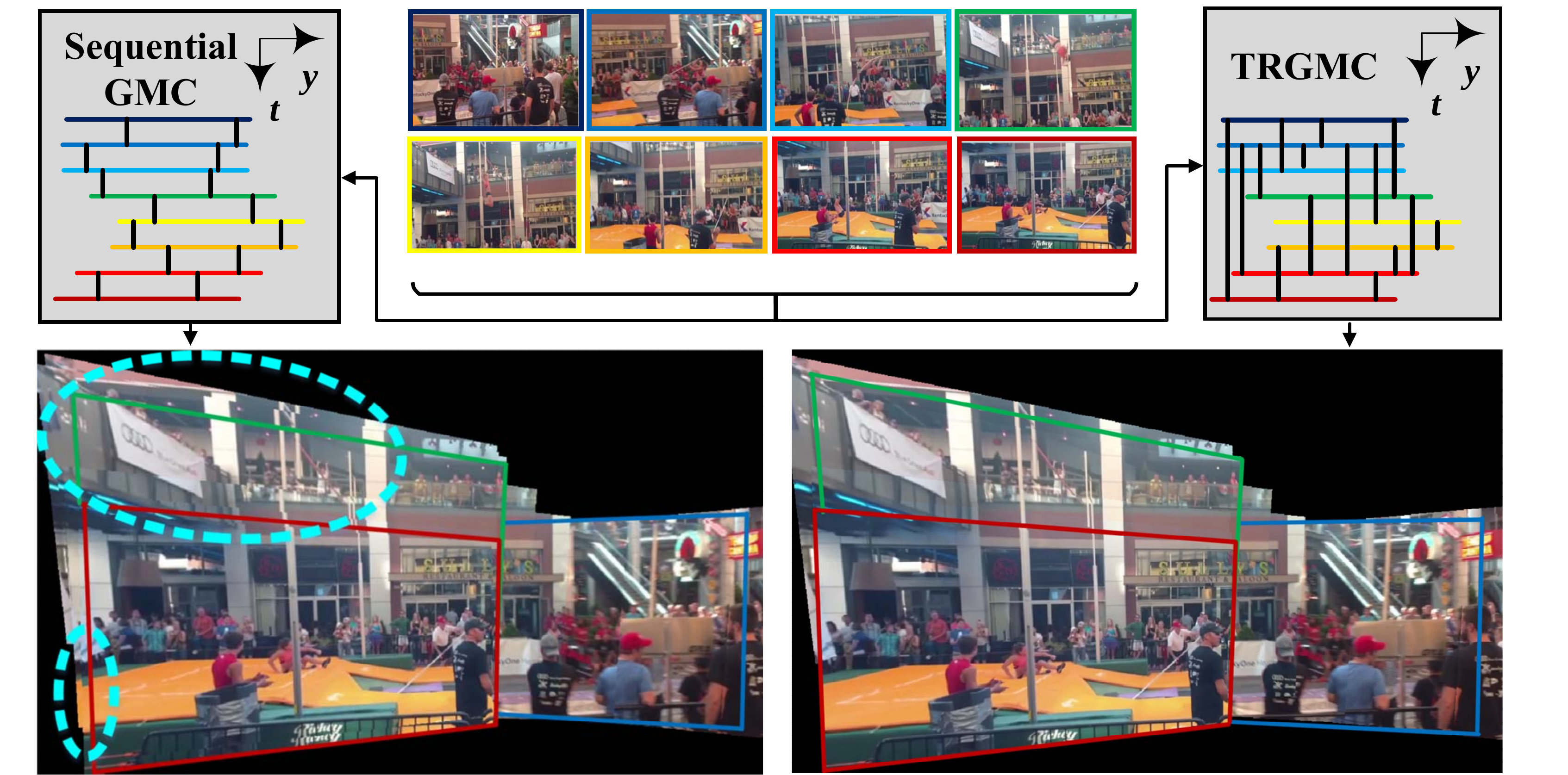} 
\vspace{-4mm}
\caption{\small Schematic diagrams of proposed TRGMC and existing sequential GMC algorithms, and resultant motion panorama for a video shot by panning the camera up and down. Background continuity breaks easily in the case of the sequential GMC~\cite{rgmc}.}
\label{fig:intro}
\figvspace
\end{figure}

%The basic idea in GMC is to register all the frames to a global coordinate so that the background appears to be static over time.
In existing GMC work~\cite{bartoli2004motion,deniz2011fast, rgmc}, frames are transformed to a global motion-compensated coordinate (GMCC), %typically the coordinates of the first input frame,
by \textit{sequentially} processing input frames.
For a pair of consecutive frames, the mapping transformation is estimated, and by accumulating the transformations, a \textit{composite} global transformation of each frame to the GMCC is obtained.
%Although creating a static background is the fundamental goal for GMC, the sequential GMC methods fail to fulfill this goal.%, by suffering from temporal drift of frames from the accurate global coordinate, either due to accumulation of error or failure in motion estimation for some of the frames. 
%For videos with enough background structure in \textit{all} the frames, as long as the camera motion is very simple (e.g. as it is for capturing a panorama image), the error in GMC by the state-of-the-art sequential procedures, such as robust global motion compensation (RGMC)~\cite{rgmc}, is negligible. 
However, the sequential processing scheme causes frequent GMC failures for multiple reasons: 
%1) In this scheme, failure in finding the correct transformation for a single frame will impact all the upcoming frames. 
%This means that for unconstrained realworld videos, even if a very short period of the input video is highly occluded by the foreground, or is affected by motion blur, the GMC will fail drastically. 
%%1) The imprecise motion compensation for one frame pair can fail the GMC of the entire video drastically.
%%This happens frequently if a frame is highly blurred or dominated by the foreground.
1) Sequential GMC is only as strong as the {\it weakest} pair of consecutive frames. A single frame with high blur or dominant foreground motion can cause the rest of the video to fail.
2) Generally, multiple planes exist in the scene.
The common assumption of a single homography will accumulate residual errors into remarkable errors.
3) Even if the error of consecutive frames is in a sub-pixel scale, due to the~\textit{multiplication} of several homography matrices, the error can be significant over time~\cite{monari2011real}. 
These problems are especially severe when processing long videos and/or the camera motion becomes more complicated. %, or zoom operation is used, accumulation of errors in the sequential schemes becomes very noticeable. 
E.g., when the camera pans to left and right repeatedly, or severe camera vibration exists, the GMC error is obvious by exhibiting discontinuity on the background (see Fig.~\ref{fig:intro} for an example).
%In this case, the resultant motion panorama loses its visual continuity, as well as the virtual static background (see Fig.~\ref{fig:intro} for an example).
%Fig.~\ref{fig:intro} illustrates such a case using the Robust GMC algorithm~\cite{rgmc}.

To address the issues of sequential GMC, we propose a temporally robust global motion compensation (TRGMC) algorithm which by~\textit{joint} alignment of input frames, estimates accurate and temporally consistent transformations to GMCC.
The result can be rendered as a motion panorama that maintains perceptual realism despite complicated camera motion (Fig.~\ref{fig:intro}).
TRGMC densely connects pairs of frames, by matching local keypoints.
Joint alignment (a.k.a.~congealing) of these frames is formulated as an optimization problem where the transformation of each frame is updated iteratively, such that for each \textit{link} interconnecting a keypoint pair, the spatial coordinates of two end points are identical.
This novel {\it keypoint-based congealing}, built upon succinct keypoint coordinates instead of high-dimensional appearance features, is the core of TRGMC.
Joint alignment not only leads to the temporal consistency of GMC, but also improves GMC stability by using redundancy of the information.
The improved stability is crucial for GMC, especially in the presence of considerable foreground motion, motion blur, non-rigid motion like water, or low-texture background.
The joint alignment scheme also provides capabilities such as coarse-to-fine alignment, i.e., alignment of the keyframes followed by non-keyframes, and appropriate weighting of keypoints matches, which cannot be naturally integrated in sequential GMC.
Our quantitative experiments reveal that TRGMC pushes the alignment error close to human performance.
%Figure~\ref{fig:intro} compares TRGMC algorithm and resultant video with sequential GMC.% a schematic diagram of TRGMC links and compares the GMC results produced by TRGMC and RGMC algorithms, for a video shot by panning the camera to the left and then back to the right.
%{\color{red} One thing we can claim is that, TRGMC is the proposed algorithm aiming for performing GMC for real-world videos. The core of TRGMC is the  keypoint-based congealing algorithm, which is a novel algorithm in  contrast to pixel-based and edge-based congealing algorithms.}

In summary, this work makes the following contributions:  
%main contribution of this paper is an algorithm for \textit{joint} alignment of the frames from a video sequence to produce a globally motion compensated video in which, no matter how complicated the camera movement is, the background appears to be static.
%Also, benefiting from redundancy of information in the joint framework, the algorithms offers improved stability.
%Thus, from a video shot by a camera with free-form movement, a video from a virtually static camera is created.
%Also, we investigate how an accurate and temporally robust GMC might improve other computer vision problems.       
\begin{itemize}
	\vspace{-2mm}
	\item An algorithm for \textit{joint} alignment of video frames is proposed to produce a globally motion compensated video where, despite the complicated camera movement and considerable foreground motion, the background appears to be static over the progression of time. \vspace{0mm}
	\item A keypoint-based congealing algorithm aligns the spatial coordinates of keypoints for an image stack.
It extends congealing applications from spatially cropped objects (faces and letters) to complex motion-rich video frames. % with diverse motion.
	\vspace{0mm}
	\item The capabilities and applications of TGRMC are demonstrated. 
Our collected video dataset, manual labels, and the code will be publicly available. 
	\vspace{0mm}
\end{itemize}
	\vspace{-4mm}
 %------------------------------------------------------------%

\Section{Prior Work}

TRGMC is related to many techniques in different aspects.
We first review them and then compare our work with existing GMC algorithms. %and contrast the most important techniques. % and contrast our work with them.
%Then, we compare our work with existing GMC algorithms.

Firstly, homography estimation from keypoint matches is crucial to many vision tasks, e.g., image stitching, registration, and GMC.
Its main challenge is the false matches due to appearance ambiguities. 
%A main challenge of homography estimation from keypoint matches is the false matches due to appearance ambiguities. 
%Robust methods are proposed to handle the outliers, such as RANSAC~\cite{ransac} and its variants~\cite{chum2003locally,mle,tordoff2005guided}. 
%Some methods also directly reject false matches~\cite{ma2014robust,li2010rejecting}.
%The hybrid methods~\cite{heask,rgmc} combines appearance similarity and keypoint matches in a probabilistic framework.
Methods are proposed to either be robust to outliers, such as RANSAC~\cite{ransac,chum2003locally,mle,tordoff2005guided} and reject false matches~\cite{ma2014robust,li2010rejecting}, or probabilistically combine appearance similarities and keypoint matches~\cite{heask,rgmc}.
%Almost all methods handle outlines using two frames.
{\it All methods estimate a homography for a frame pair.}
In contrast, we jointly estimate homographies of {\it all frames} to a global coordinate, which leverages the redundant background matches over time to better handle outliers.
%However, in TRGMC, instead of direct calculation of homography transformation for each   pair of frame, we jointly optimize the set of homographies which map the set of input frames into a global coordinate, such that the keypoints over a wide range of temporal distance are aligned well. 
 
%The obtained transformation to GMCC may be used to render input video to a motion panorama.
Image stitching (IS) and panoramic image mosaicing share similarity with GMC. %motion panorama. 
IS aims to minimize the  distortions and  ghosting artifact in the overlap region. % distortions and after stitching images.
Recent works focus on different challenges, e.g., multi-plane scenes ~\cite{szpak2015robust, zuliani2005multiransac, toldo2008robust, ma2014mixture, uemura2008feature,gao2011constructing}, the parallax issue~\cite{lin2011smoothly,zaragoza2014projective,lin2015adaptive}, and motion blur~\cite{li2010generating}.
%Many works utilize multiple homographies, instead of a~\textit{single} homography, due to existence of multiple scene planes~\cite{szpak2015robust,zuliani2005multiransac, toldo2008robust, ma2014mixture, uemura2008feature,gao2011constructing}.
%Some recent works focus also on the parallax issue
%%, by using a hybrid model that uses homograhy for non-overlapping parallax-free regions and allow some local non-projective deviation to account for parallax and avoid stitching artifacts
%~\cite{lin2011smoothly,zaragoza2014projective,lin2015adaptive}.
%Li~\etal~\cite{li2010generating} generate panoramas from motion-blurred videos.
In these works, input images have much less overlap than GMC.
On the other hand, video mosaicing takes in a video which raster scans a wide angle~\textit{static} scene, and produces a single~\textit{static} panoramic image~\cite{sawhney1998robust,shum1998construction,sakamoto2006homography}. % of the scene.  
When the camera path forms a $2$D scan~\cite{sawhney1998robust} or a $360^{\circ}$ rotation~\cite{sakamoto2006homography}, global refinement is performed via bundle adjustment (BA)~\cite{triggs1999bundle}, which ensures an artifact-free panoramic image.
%Also, the camera optical center is normally assumed to be fixed, i.e., the camera has no considerable transformation.
%Assuming a static scene, many of these works rely on bundle adjustment~\cite{triggs1999bundle} for global refinment. % of estimations.%estimates 3D location of sparse points and camera pose and deals mainly with~\textit{static} content.
%However, the pixel-based global refinements employed in~\cite{sawhney1998robust} fall short in terms of efficiency and robustness to foreground motion in a dynamic scene. 
Although a byproduct of TRGMC is a similar static reconstruction of the scene, TRGMC focuses on efficient generation of an appealing video,  %, where background consistently appears static for visual perception (in contrast to an image), in which background consistently appears static, 
for a~\textit{highly dynamic} scene. %, and is usable for further video processing
%The important feature of such a video is that the only apparent motion in the video will rise from foreground motion.
%So, despite similarities, 
While one may use BA to estimate camera pose and then transformation between frames, our experiments reveal that BA is not reliable for videos with  foreground motion and is far less efficient than TRGMC. 
%Further, BA estimates 3D location of keypoints while TRGMC needs 2D registration. 
%Thus, by using BA, a harder problem needs to be solved which is unnecessary for the purpose of global motion compensation.
Hence, image/video mosaicing and GMC have different application scenarios and challenges.

%Instead of creating a single static mosaic image, GMC generates a video sequence for which the background remains static over time. 
%{\color{red} Also, handling considerable foreground motion differentiates TRGMC from these works.}

Another related topic is the panoramic video~\cite{el2011improved, perazzi2015panoramic, zeng2009depth, jiang2015video, ibrahim2012automatic}.
For instance, Perazzi~\etal~\cite{perazzi2015panoramic} create a panoramic video from an array of stationary cameras by generalizing parallax-tolerant image stitching to video stitching.
%In~\cite{zeng2009depth} also a method for fast stitching of videos from multiple cameras is proposed, which 
%The fast video stitching in~\cite{zeng2009depth} 
%can handle proper stitching of objects at varying depths.
%Jiang and Gu~\cite{jiang2015video} propose an algorithm for stitching multiple video streams into a single panoramic video with spatial-temporal content-preserving warping.
% In this work, for alignment of video frames, a spatial-temporal local warping is proposed, which locally aligns frames from different videos while maintaining the temporal consistency.
While these works focus on stitching~\textit{multiple} synchronized videos, GMC creates a motion panorama from a \textit{single non-stationary} camera.
Unlike GMC, video panoramas do not require the resultant video to have a stationary background.
%The significant difference between the video panoramas and GMC is the requirement of stationary background for the video created by GMC.

Video stabilization (VS) is a closely related but different problem.
TRGMC can be re-purposed for VS, but not vice versa, due to the accuracy requirement. 
Given the accurate mapping to a global coordinate using TRGMC, VS would mainly amount to cropping out a smooth sequence of frames and handling rendering issues such as parallax.
Among different categories of VS, $2$D VS methods calculate consecutive warping between the frames and have similarities with~\textit{sequential} GMC, but any estimation error will not cause severe degradation in VS as long as it is smoothed. 
While TRGMC targets~\textit{long-term staticness of the background}, VS mainly cares about~\textit{smoothing} of camera motion, not~\textit{removing} it. 
In other words, TRGMC imposes a stronger constraint on the result. % which is background staticness by complete camera motion removal  in comparison to VS which deals with camera motion smoothing.
This strict requirement differentiates TRGMC also from Re-Cinematography~\cite{gleicher2008re}.
%Also, large occlusion by the foreground may result in VS failure, however TRGMC handles this challenge by utilizing redundancy of background information in the joint alignment scheme.

Congealing aims to jointly align a stack of images from one object class, e.g., faces and letters~\cite{learned2006data,Liu2009b}.
Congealing iteratively updates the transformations of all images such that the entropy~\cite{learned2006data} or Sum of Squared Differences (SSD)~\cite{cox2008least} of the images, is minimized.  
%Recently, many extensions of congealing are proposed~\cite{huang2012learning, lankinen2011local, lucey2013fourier, cox2009least,shokrollahi2015unsupervised}.
%computing the empirical distribution defined by a set of images,
%then for each image, choosing a transformation that reduces the entropy of the distribution field. 
%The rational for using such approach is that total entropy for an ensemble of aligned images is less than the total entropy in an ensemble of misaligned images~\cite{cox2008least}.
However, despite many extensions of congealing~\cite{huang2012learning, lankinen2011local, lucey2013fourier, cox2009least,shokrollahi2015unsupervised}, almost all prior work define the energy based on the {\it appearance features} of two images.
Our experiments on GMC show that appearance-based congealing is inefficient and sensitive to initialization and foreground motion. %, especially when the stack size increases.
Therefore, we propose a novel keypoint-based congealing algorithm minimizing the SSD of corresponding~\textit{keypoint coordinates}. %, instead of appearance features. %to gain considerable efficiency enhancement and robustness to initialization.%, our formulation relies on coordinates of keypoints.
Further, most prior works apply to a spatially cropped object such as faces, while we deal with complex video frames with dynamic foreground and moving background, at a higher spatial-temporal resolution.
Note that~\cite{lankinen2011local} uses a heuristic local feature based algorithm to rigidly align object class images.
In contrast we formulate the joint alignment of keypoints as an optimization problem and solve it in a principal way.

There are a few existing sequential GMC works, % rely on the sequential scheme.
where the main problem is to accurately estimate a homography transformation between consecutive frames, given challenges such as appearance ambiguities, multi-plane scene, and dominant foreground~\cite{rgmc, deniz2011fast,bartoli2002video}.
Bartoli~\etal~\cite{bartoli2004motion} first estimate an approximate $4$-degree-of-freedom homography, 
%At this stage, assuming that camera %is mounted on the tripod and 
%does not have any rotation around the optical axis, and also has almost fixed focal length, the degrees of freedom of homography  is limited to $4$ . 
and then refine it.
%In~\cite{bartoli2002video}, dynamic scenes where the camera is only rotating are considered.
%Deniz~\etal~\cite{deniz2011fast}  estimate only $2$D translation by correlating the $x$ and $y$ projections of pixels between frames. 
Sakamoto~\etal~\cite{sakamoto2006homography} generate a $360^\circ$ panorama from an image sequence.
%Assuming that multiplication of all consecutive homographies results in the {\color{red} identity} mapping, and homography has only $5$ degrees of freedom, %(the camera rotation matrix has $3$ degrees of freedom, to which are added the focal lengths before and after the camera rotation)
Assuming a $5$-degree-of-freedom homography, all the homographies are optimized jointly to prevent error accumulation.   
In contrast, TRGMC employs an $8$-degree-of-freedom homography.
Although using homography in the case of considerable camera translation and large depth variation results in parallax artifacts, using a higher degrees-of-freedom homography than prior works allows TRGMC to better handle camera panning, zooming, and translation.
 %, and is thus less restricted in comparison to previous works. 
%Using a method similar to RANSAC, at each iteration a subset of projections of some horizontal and vertical stripes of the image are utilized in the correlation operation, resulting into some robustness to presence of motion of small foreground.
Safdarnejad~\etal~\cite{rgmc} incorporate edge matching into a probabilistic framework that scores candidate homographies.
%To suppress effect of foreground, motion vectors obtained from matching the keypoints   are clustered and each cluster is analyzed separately. 
%Since background motion vectors are more consistent in pattern in comparison to those of foreground, clusters with best matching scores are assumed to belong to the background region.
%Thus, these clusters are merged back together to form the set of suitable matches, not affected by the foreground presence, to be used for the estimation of final homography.
Although~\cite{rgmc, deniz2011fast} improve the robustness to foreground,  %and ~\cite{rgmc} targets a more accurate homography by utilizing a sophisticated probabilistic model
error accumulation and failure in a single frame pair still deteriorate the overall performance.
Thus, TRGMC targets robustness of the GMC in terms of both the presence of foreground and long-term consistency by 
%utilizing the redundancy of information in 
joint alignment of frames.

%For TRGMC, we do not impose any constraint regarding the composite homography and the degrees of freedom.

\begin{figure*} [t!]
\begin{center}
\includegraphics[trim=.5cm 4.5cm 0cm 25.7cm,clip,width=\textwidth]{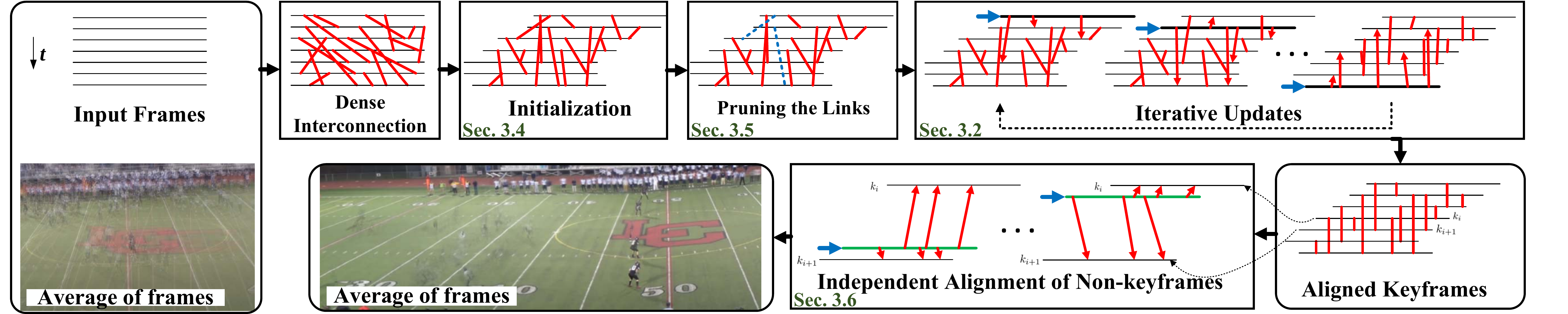}
\end{center} \vspace{-6mm}
\caption{\small Flowchart of the TRGMC algorithm.}
\label{fig:schematic}
\figvspace
\end{figure*}

\Section{Proposed TRGMC Algorithm}
\label{sec:proposed}

%-----Requiring good initialization is a drawback in congealing framework, but our proposed method does not suffer from this challenge. 
%Since in our case the stack of frames to be aligned are from the same scene, it is straightforward to find an approximate initialization, as will be discussed in sec.~\ref{sec:proposed}.

%Our proposed method formulates the GMC problem as the joint alignment of frames, resulting in temporally robust GMC. % (TRGMC).
%For the joint alignment problem to be feasible, we build our method upon the keypoint-based, rather than appearace-based~\cite{heask} or edge-based alignment~\cite{rgmc}.
%Requiring good initialization is a drawback in congealing which is also alleviated by using the keypoints.
The core of TRGMC is the novel keypoint-based congealing algorithm.
Our method relies on densely interconnecting % spatially overlapping region of 
the input frames, regardless of their temporal offset, by matching the detected SURF~\cite{surf} keypoints at each frame.
We refer to these connections, shown in Fig.~\ref{fig:schematic}, as \textit{links}.
%In this figure, 
Frames are initialized to their approximate spatial location by only $2$D translation (Sec.~\ref{sec:init}).
We rectify the keypoints such that majority of the links have end points on the background region.
Then the congealing applies appropriate transformation to each frame and the links connected to it, such that the spatial coordinates of the end-points of each link are as similar as possible. 
In Fig.~\ref{fig:schematic}, this translates to having the links as parallel to the $t-$axis as possible.

For efficiency and robustness, TRGMC processes an input video in two stages. 
%First, we select a stack of keyframes and jointly align this stack.
%Then, we freeze the transformation of keyframes and perform the keypoint-based congealing for each remaining frame, using \textit{only} its two encompassing keyframes. 
%In the rest of this section, we present various aspects of our algorithm.
Stage one selects and jointly aligns a set of keyframes.
The keyframes are frozen, and then stage two aligns each remaining frame to its two encompassing keyframes. % with keypoint-based congealing.
The remainder of this section presents the details of the algorithm.

%In the following sections, we will discuss the details about joint alignment of a stack of frames, selection of keyframes, alignment of the non-keyframes, initialization of the problem, and pruning the keypoints such that the majority of keypoints correspond to the background.
%We will also discuss how existence of foreground motion is handled in TRGMC.

\SubSection{Formulation of keypoint-based congealing}
\label{sec:jointalignment}

Given a stack of $N$ frames $\{\mathbf{I}^{(i)}\}$, with indices $i\in\mathbb{K}=\{k_1, ..., k_N\}$, the keypoint-based congealing is formulated as an optimization problem,
\eqnvspace
\begin{equation}
\mymin_{\{ \p_i \}}{\epsilon = \sum_{i \in \mathbb{K}}{[\E_i(\p_i)]^\T \Omega^{(i)} [\E_i(\p_i)]}},
\label{equ:initialCost}
\eqnvspace\vspace{-2mm}
\end{equation}
where $\p_i$ is the transformation parameter from frame $i$ to GMCC, $\E_i(\p_i)$ collects the pair-wise alignment errors of frame $i$ relative to all the other frames in the stack, and $\Omega^{(i)}$ is a weight matrix.

\begin{figure} [t!]
\centering
%\begin{tabular}{p{3.5cm}p{3.3cm}}
\begin{tabular}{c}
\includegraphics[trim=.3cm 0cm .2cm 1.1cm,clip,height=2.7cm]{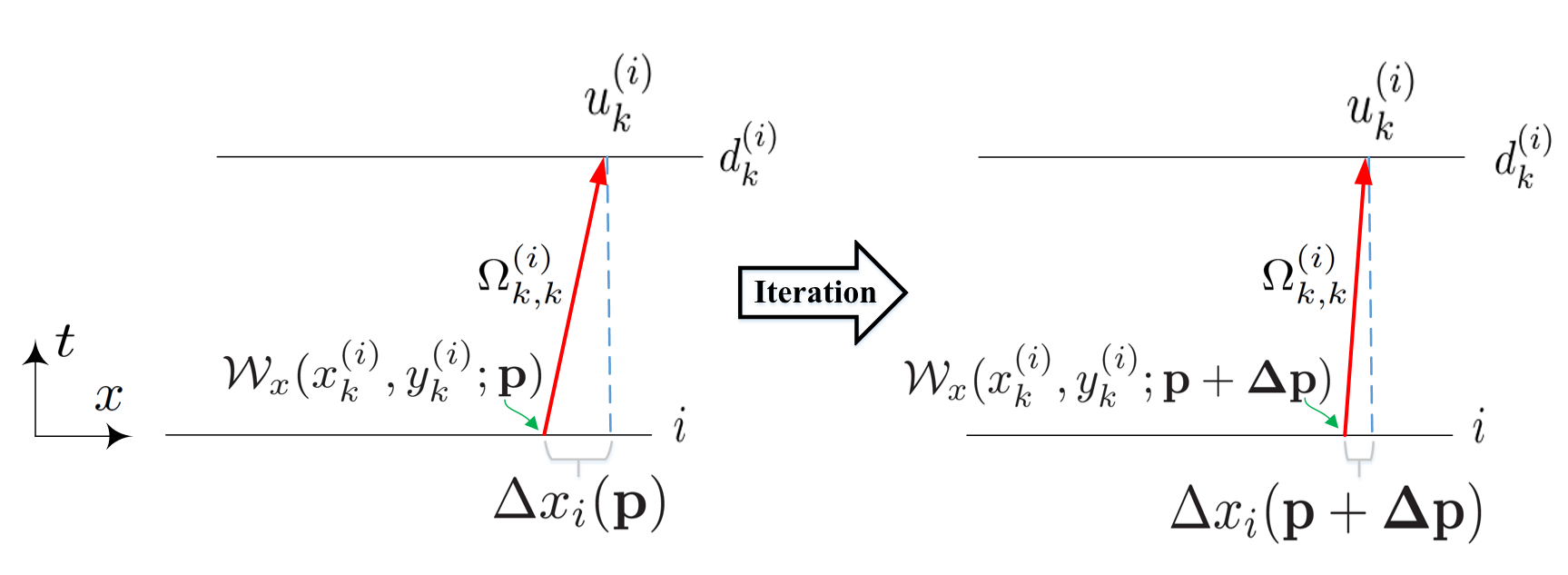} 
 \vspace{-1mm}\\
%\centering \small (a) &  \centering \small (b) \\
\end{tabular}
\vspace{-4mm}
\caption{\small The notation used in TRGMC.}
\vspace{-2mm}
\label{fig:notation}
\figvspace
\end{figure}

%For the case of keypoint based alignment of frames, we define the error of each frame relative to the placement of other frames, 
We define the alignment error of %two frames 
frame $i$ %relative to other frames 
as the SSD between the spatial coordinates of the endpoints of all links connecting frame $i$ to the other frames, instead of the SSD of appearance~\cite{cox2008least}. 
Specifically, as shown in Fig.~\ref{fig:notation}, we denote coordinates of the start and the end point of each link $k$ connecting frame $i$ to the frame $d_k^{(i)} \in \mathbb{K}\backslash \{i\}$ as $(x_k^{(i)}, y_k^{(i)})$ and $(u_k^{(i)}, v_k^{(i)})$, respectively. 
For simplicity, we omit the frame index $i$ in $\p_i$.
Thus, the error $\E_i(\p)$ is defined as, 
\eqnvspace
\begin{equation}
\E_i(\p) =  [\DX_i(\p)^\T, \DY_i(\p)^\T]^\T, %[\DX_i(p), \DY_i(p)],
\vspace{-3mm}
\end{equation}
where %in which, 
\begin{equation}
%\DX_i(\p) = [\mathcal{W}_x(x_k^{(i)}, y_k^{(i)}; \p)-u_k^{(i)}]; \ \  k \in \{1,...,N_i\}; \ \ 
\DX_i(\p) = \wxi - \mathbf{u}^{(i)},
%\eqnvspace
%\end{equation}
%\begin{equation}
%\DY_i(\p) = [ \mathcal{W}_y(x_k^{(i)}, y_k^{(i)}; \p)-v_k^{(i)}]; \ \ k \in \{1,...,N_i\}; \ \ 
\quad \DY_i(\p) = \wyi - \mathbf{v}^{(i)},
\eqnvspace
\end{equation}
are the errors in $x-$ and $y-$ axes.
The vectors $\wxi=[\mathcal{W}_x(x_k^{(i)}, y_k^{(i)}; \p)]$ and $\wyi=[\mathcal{W}_y(x_k^{(i)}, y_k^{(i)}; \p)]$ denote the $x$ and $y-$ coordinates of $(x_k^{(i)}, y_k^{(i)})$ warped by the parameter $\p$, respectively. 
The vectors $\mathbf{u}^{(i)}=[u_k^{(i)}]$ and $\mathbf{v}^{(i)}=[v_k^{(i)}]$ denote the coordinates of the end points.
Similarly, the vectors $\x^{(i)}=[x_k^{(i)}]$ and $\y^{(i)}=[y_k^{(i)}]$ denote the coordinates of the start points.
If $N_i$ links emanate from frame $i$, $\E_i$ is a $2N_i-$dim vector.
$\Omega^{(i)}$ is a diagonal matrix of size $2N_i \times 2N_i$ which assigns a weight to each element in $\E_i$.
The parameter $\p$ has $2$, $6$, or $8$ elements for the cases of $2$D translation, affine transformation, or homography, respectively. 
In this paper, we focus on homography transformation which is a projective warp model, parameterized as,
\vspace{-4mm}
\begin{equation}
\begin{bmatrix}
\mathcal{W}_x(x_k^{(i)}, y_k^{(i)}; \p) \\
\mathcal{W}_y(x_k^{(i)}, y_k^{(i)}; \p) \\
1
\end{bmatrix} =% \p
%\begin{bmatrix}
%x_k^{(i)} \\
%y_k^{(i)} \\
%1
%\end{bmatrix}=
\overbrace{
\begin{bmatrix}
p_1 & p_2 & p_3 \\
p_4 & p_5 & p_6 \\
p_7 & p_8 & 1
\end{bmatrix}
}^{\p}
\begin{bmatrix}
x_k^{(i)} \\
y_k^{(i)} \\
1
\end{bmatrix}.
\vspace{-3mm}
\end{equation}

Although the homography model assumes the planar scene and this assumption may be violated in real world~\cite{zaragoza2014projective}, we identify the problem of temporal robustness to be more fundamental for GMC than the inaccuracies due to a \textit{single} homography.
Also, videos for GMC are generally swiped through the scene with high overlap, thus the discontinuity resulted from this assumption is minor.

\SubSection{Optimization solution}
\label{sec:jointalignmentsolution}
%Eqn.
Equation~\ref{equ:initialCost} is a non-linear optimization problem and difficult to minimize. 
Following~\cite{cox2008least}, we linearize this equation by taking the first-order Taylor expansion around $\p$.
Starting from an initial $\p$, the goal is to estimate $\Delta \p$ by,

\eqnvspace
\begin{equation}
%\argmin_{\Delta \p}{[\E_i(\p)+\frac{{\partial \E_i(\p)}^T}{\partial \p} \Delta \p]^2} + \gamma \parallel {\Delta \p} \parallel^2_\mathcal{I}.
\argmin_{\Delta \p}{[\E_i(\p)+\frac{{\partial \E_i(\p)}}{\partial \p} \Delta \p]^\T \Omega^{(i)} [\E_i(\p)+\frac{{\partial \E_i(\p)}}{\partial \p} \Delta \p]} 
+ \gamma {\Delta \p}^\T \mathbf{\mathcal{I}} \Delta \p,
\label{eqn_delpObj}
\eqnvspace
\end{equation}
%in which $\parallel {\Delta \p} \parallel^2_\mathcal{I}$ is added as a regularization term, defined as
where ${\Delta \p}^\T \mathbf{\mathcal{I}} \Delta \p$ is a regularization term, with a positive constant $\gamma$ setting the trade-off.
We observe that without this regularization, parameter estimation may lead to distortion of the frames.
The indicator matrix $\mathbf{\mathcal{I}}$ is a diagonal matrix specifying which elements of $\Delta \p$ need a constraint. 
%For instance, when there is not an accurate parameter $\p$ which maps frame $i$ to the other frames, or in case that all the keypoints are concentrated on a small spatial region, this distortion is more probable. 
%However, since in the GMC problem all the frames come from a single video sequence, we wish to warp the frame as little as possible, and yet, align them well.
%Thus we regularize the cost function as,
We use $\mathbf{\mathcal{I}} = diag([1, 1, 0, 1, 1, 0, 1, 1])$ to specify that there is no constraint on the translation parameters of the homography, but the rest of parameters should remain small.

%By taking the first-order derivative of Eqn.~\ref{eqn_delpObj} and setting it to zero, the solution for $\Delta \p$ is given by,
By setting the first-order derivative of Eqn.~\ref{eqn_delpObj} to zero, the solution for $\Delta \p$ is,
\vspace{-3mm}
\begin{equation}
\vspace{-3mm}
\Delta \p = \mathbf{H}^{-1}_R  { \frac{{\partial \E_i(\p)}^\T}{\partial \p}   \Omega^{(i)} \E_i(\p)},
\label{eqn_solution}
\eqnvspace
\end{equation}
%in which,
\vspace{1mm}
\begin{equation}
\eqnvspace
\mathbf{H}_R =  { {\frac{\partial \E_i(\p)}{\partial \p}}^\T  \Omega^{(i)}  \frac{\partial \E_i(\p)}{\partial \p} }+  \gamma \mathbf{\mathcal{I}}.
\eqnvspace
\end{equation}

Using the chain rule, we have $\frac{{\partial \E_i(\p)}}{\partial \p}=\frac{{\partial \E_i(\p)}}{\partial \mathcal{W}} \frac{{\partial \mathcal{W}}}{\partial \p}$.  
Knowing that the mapping has two components as $\mathcal{W}=(\mathcal{W}_x, \mathcal{W}_y)$, and the first half of $\E_i$ only contains $x$ components and the rest only $y$ components, we have,
\eqnvspace
\begin{equation}
\frac{{\partial \E_i(\p)}}{\partial \mathcal{W}}  = \left[ \begin{smallmatrix} \Nione & \Nizero \\ \Nizero & \Nione \end{smallmatrix} \right],
\eqnvspace
\end{equation}
where $\Nione$ and $\Nizero$ are $N_i-$dim vectors with all element being $1$ and $0$, respectively.
For homography transformation, $\frac{{\partial \mathcal{W}}}{\partial \p}=\frac {\partial (\mathcal{W}_x, \mathcal{W}_y)}{\partial (p_1, p_2, p_3, p_4, p_5, p_6, p_7, p_8)}$ is given by,
\vspace{-1mm}
\begin{equation}
\frac{{\partial \mathcal{W}}}{\partial \p}  = 
 \left[ \begin{smallmatrix} \wxi &\wyi&\Nione&\Nizero&\Nizero&\Nizero&-\mathbf{u}^{(i)}\wxi&-\mathbf{u}^{(i)}\wyi\\ 
\Nizero&\Nizero&\Nizero&\wxi &\wyi&\Nione&-\mathbf{v}^{(i)}\wxi&-\mathbf{v}^{(i)}\wyi
 \end{smallmatrix} \right].
\eqnvspace
\end{equation}

At each iteration, and for each frame $i$, $\Delta \p$ is calculated and the start points of all the links emanating from frame $i$ are updated accordingly.
Similarly, for all links with end points on frame $i$, the end point coordinates are updated.
\footnote{
%\scriptsize
In algorithm implementation, it is important to store the original coordinates of the detected keypoints and apply the~\textit{composite} transformations accumulated in all the iterations to update the coordinates of the start and end points of the links. 
Otherwise, accumulation of numerical errors will harm the performance.} %if the links from the previous iteration are updated with the transformation at the current iteration, 

We use the SURF~\cite{surf} algorithm for keypoint detection %. 
%We reduce the Fast-Hessian keypoint detection threshold, $\tau_s$, from the typical value of $1,000$ to $200$, so that sufficient keypoints are detected, even in low-texture background.
with a low detection threshold, $\tau_s = 200$, to ensure sufficient keypoints are detected even for low-texture backgrounds.
We use the nearest-neighbor ratio method~\cite{lowe2004distinctive} to match keypoints and form links between each pair of keyframes.

\Paragraph{Keyframe selection}
We select keyframes at a constant step of $\Delta f$, i.e., from every $\Delta f$ frames, only one is selected.
Based on the experimental results, as a trade-off between accuracy and efficiency, we use $\Delta f=10$ in TRGMC.

\SubSection{Weight assignment}
We have defined all parameters in the problem formulation, except the weights of links, $\Omega^{(i)}$.
We consider two factors in setting $\Omega^{(i)}$.
Firstly, the keypoints detected at larger scales are more likely to be from background matches, since they cover coarser information and larger image patches.
Thus, to be robust to foreground, the early iterations should emphasize links from larger-scale keypoints, which forms a coarse-to-fine alignment.
% FIXME, coase to fine is not the reason, be robust to background is the reason?
We normalize the scales of all keypoints such that the maximum is $1$, and denote the minimum of the normalized scales of the two keypoints comprising the link $k$ as $s_k$. 
Then, $\Omega^{(i)}_{k,k}$ is set proportional to $s_k$.

Secondly, for each frame $i$, the links may be made either to all the previous frames, denoted as \textit{backward} scheme, or both the previous and upcoming frames, denoted as \textit{backward-forward} scheme. 
The former is for potential real time application, whereas the latter for offline video processing. 
These schemes are implemented by assigning different weights to backward and forward links,
\vspace{-3mm}
\begin{equation}
  \Omega^{(i)}_{k,k}=\begin{cases}
               \  (\beta . s_k)^{r^q}; \  \mbox{if}  \ \  d_k^{(i)}<i \  \  \       \mbox{\small(Backward \ links)} \\
               \  (\alpha . s_k)^{r^q}; \  \mbox{if}  \ \  d_k^{(i)}>i \  \  \     \mbox{\small(Forward \ links)}\\
            \end{cases}
\label{eqn:weights1}
\eqnvspace
\end{equation}
where $0<\alpha,\beta<1$, $q$ is the iteration index, and $0<r<1$ is the rate of change of the weights.
Note that the alignment errors in $x$ and $y-$axes have the same weights, i.e., $\Omega^{(i)}_{k+N_i,k+N_i}=  \Omega^{(i)}_{k,k}$.
After a few iterations, the weights of all the links will be restored to $1$.
In the backward scheme, we set $\alpha = 0$.

\definecolor{cyan}{RGB}{70,130,180}
\iffalse
\begin{figure} [t!]
\centering
\begin{tabular}{p{0.01cm}p{3.1cm}p{0.01cm}p{3cm}}
\small (a) &
\includegraphics[trim=17cm 29.2cm 38cm 10.3cm,clip,height=1.7cm]{images/redundancy_1.pdf} &
\small (b) &
\includegraphics[trim=21.5cm 29.2cm 33cm 10.3cm,clip,height=1.7cm]{images/redundancy_2.pdf} \\
\end{tabular}
\caption{\small Comparison of the ratios of \textbf{{\color{red}background}}-\textbf{{\color{cyan}foreground}} matches for (a) sequential GMC and (b) TRGMC.}
\label{fig:redundancy}
\figvspace
\end{figure}
\fi

\begin{figure} [t!]
\centering
%\includegraphics[trim=0cm 0cm 0cm 0cm,clip,height=2cm]{images/steps.pdf} 
%\vspace{-2mm}
%\caption{\small Error and efficiency vs.~the keyframe selection step, $\Delta f$.}
%\label{fig:tradeoff}

\begin{minipage}[t]{0.45\linewidth}
\begin{tabular}{p{0.01cm}p{2.6cm}p{0.01cm}p{2cm}}
	\small (a) &
	\includegraphics[trim=17cm 29.2cm 38cm 10.3cm,clip,height=1.7cm]{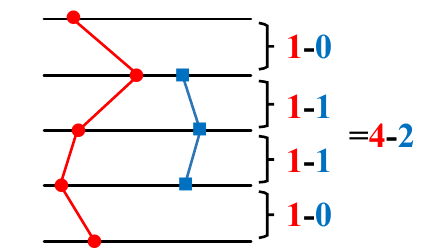} &
	\small (b) &
	\includegraphics[trim=21.5cm 29.2cm 33cm 10.3cm,clip,height=1.7cm]{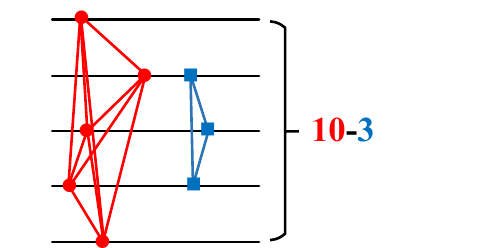} \\
	\end{tabular}
	\vspace{-1.5mm}
	\caption{\small Comparison of the ratios of \textbf{{\color{red}background}}-\textbf{{\color{cyan}foreground}} matches for (a) sequential GMC and (b) TRGMC.}
	\label{fig:redundancy}
\end{minipage}%
\hfill%
\begin{minipage}[t]{0.50\linewidth}

\centering
%\begin{tabular}{llll}
\begin{tabular}{p{0.4cm}p{2.5cm}p{0.4cm}p{2.6cm}}
\small (a) &
\includegraphics[trim=0cm 0cm 0cm 0cm,clip,width=2.5cm]{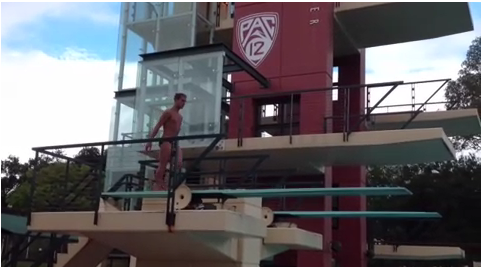} &
\small (b) &
\includegraphics[trim=0cm 0cm 0cm 0cm,clip,width=2.5cm]{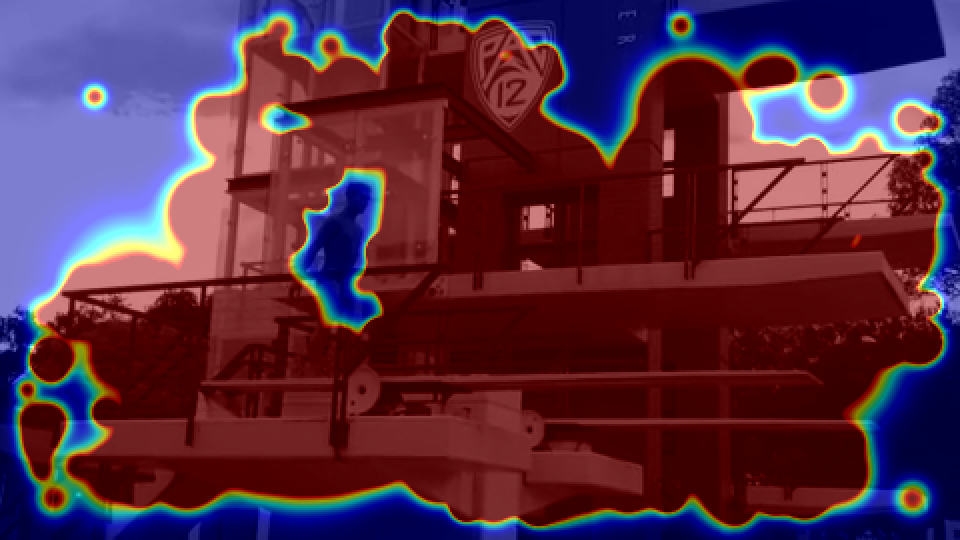} \\
\end{tabular}
\caption{\small (a) The input frame, (b) the reliability map, with the red color showing higher reliability.}
\label{fig:reliabilitymap}
\figvspace

\end{minipage} 
\figvspace
\end{figure}

\SubSection{Initialization}
\label{sec:init}
Initialization speeds up the alignment and decreases the false keypoint matches. 
The objective is to roughly place each frame at the appropriate coordinates in the GMCC.
%For TRGMC, we only initialize the transformation for each frame to the translation needed in $x-$ and $y-$ direction to roughly align the frames. 
For initialization, we align the frames based only on rough estimation of translation without considering rotation, skew, or scale.
%Two simple methods can be used in this regard.  
%Similar to Deniz \etal~\cite{deniz2011fast} work, one can correlate the intensity projection of consecutive frames onto $x-$ and $y-$ axes to find the translation. % between consecutive frames.%, and use this to find the composite translation for each frame to roughly align it to the first frame.
%Also,} 
We use the average of the motion vectors in matching two consecutive frames as the translation. % and find the composite translation.
%{\color{blue}We use the latter in TRGMC.}
Using this simple initializiation, even if the camera has  in-plane rotation, estimated 2D translations are zero, which is indeed correct and does not cause any problem for TRGMC.
Given the estimated translation, approximate overlap area of each pair of frames is calculated, and only the keypoints inside the overlap area are matched, reducing number of false matches due to appearance ambiguities. 
%This further reduces the number of false matches due to appearance ambiguities. 

\SubSection{Outlier handling}
\label{sec:robustness}
%\subsection{Pruning the links}
%Generally, a large number of keypoint matches are not background matches. %; they are outliers and need to be handled in GMC.
%Many are false matches caused by the employed low detection threshold and appearance ambiguities. % e.g., repetitive background patterns.  %and  is used.%for the sake of low-texture backgrounds.
%Also, some of the keypoint matches reside on the foreground, which are not necessarily consistent with the camera motion.

Links may become outliers for two reasons: 
(i) the keypoints reside on foreground objects not consistent with camera motion;  
(ii) false links between different physical locations are caused by the low detection threshold and similar appearances.

In order to prune the outliers, we assume that the motion vectors of background matches, i.e., background links, have consistent and smooth patterns, caused by camera motion such as pan, zoom, tilt, whereas, the outlier links will exhibit arbitrary pattern, inconsistent with the background pattern.
%The distinct and smooth patterns of these motion vectors, generated by camera motion%pan, zoom, tilt, and in-plane rotation, and also camera translation
%, is utilized to prune non-background matches.
% FIXME  not clear
Specifically, we use Ma~\etal~\cite{ma2014robust} method to prune outlier links by imposing a smoothness constraint on the motion vector field\footnote{We use the implementation provided by the authors and default parameters.}.
%This problem is formulated in a mixture model by defining latent variables for pruning non-background matches by imposing a smoothness constraint on the vector field.
This method outperforms RANSAC if the set of keypoint matches contains a large proportion of outliers. %~\cite{ma2014robust}.
Since keyframes have larger relative time difference than consecutive frames, the foreground motion is accentuated and more distinguishable from camera motion. %, especially for keyframes further apart. 
This helps with better pruning of the foreground links.
At each stage that the keypoints from a pair of frames are matched to form the links, we perform the pruning. 
 %far away from each other.

%On the other hand, TRGMC is formulated such that the details of each image in the stack are used for alignment, rather than only relying on the average image of the stack, as in entropy-based congealing.
%Thus, redundancy of background information over time brings in robustness to false matches and existence of foreground.
%On the other hand, 
Congealing of an image stack also increases the proportion of background matches over the outliers - another way to suppress outliers. 
%The reason is that 
The keypoints on background are more likely to form longer range matches than the foreground ones, due to non-rigid foreground motion.  
Hence, when $\binom {N}{2}$ combinatorial pairs of frames are interconnected, there are a lot more background matches (Fig.~\ref{fig:redundancy}). 
%Also, motion pattern of the foreground is generally inconsistent and non-smooth, both due to existence of different objects/people in the scene and motion of articulated body.
%Pruning the links based on the smoothness of the motion vector field, adds more robustness to the algorithm. 

%In summary, the redundancy of information along with pruning the links, help the alignment framework to align the frames based on the background content, not the foreground motion.
%In fact, foreground alignment will be more costly in terms of the cost function, in comparison to aligning the background area, specifically if background region has more or comparable number of links in comparison to the foreground, which is the case after pruning the links.

\iffalse
\begin{figure} [t!]
\centering
%\begin{tabular}{llll}
\begin{tabular}{p{0.5cm}p{3.5cm}p{0.5cm}p{3.4cm}}
\small (a) &
\includegraphics[trim=0cm 0cm 0cm 0cm,clip,width=3.5cm]{images/reliabilityMapInput.png} &
\small (b) &
\includegraphics[trim=0cm 0cm 0cm 0cm,clip,width=3.5cm]{images/reliabilityMap.png} \\
\end{tabular}
\caption{\small (a) The input frame, (b) the reliability map, with the red color showing higher reliability.}
\label{fig:reliabilitymap}
\figvspace
\end{figure}
\fi

\SubSection{Alignment of non-keyframes}
\label{sec:nonkeyframes}
The keyframes alignment provides a set of temporally consistent motion compensated frames, which are the basis for aligning non-keyframes.
%In this work 
We refer to keyframes and non-keyframes with superscripts $i$ and $j$, respectively.
For a non-keyframe $j$ between the keyframes $k_{i}$ and $k_{i+1}$, its alignment is a special case of Eqn.~\ref{equ:initialCost}, with indices $\mathbb{K}=\{j\}$, and the destination of the links $d_k^{(j)} \in \{ k_i, k_{i+1} \}$, i.e., only $\p_j$ of frame $j$ is updated while the keyframes remain fixed. % links are made between frame $j$ and $k_{i}$, and also $j$ and $k_{i+1}$.%, with equal weight of $1$.  
%Iteratively, \textit{only} frame $j$ is aligned and the links emanating from it are updated, as discussed in Sec.~\ref{sec:jointalignmentsolution}.
Each non-keyframe between keyframes $k_{i}$ and $k_{i+1}$ is aligned independently.

However, given the small time offset between $j$ and $d_k^{(j)}$,
%between each non-keyframe and encapsulating keyframes, 
 the observed foreground motion may be hard to discern. 
Also, frame $j$ is linked only to two keyframes, thus %in contrast to the case of the keyframes, here redundancy of background information and inconsistency of the foreground motion cannot boost the robustness of the algorithm to existence of the foreground. 
there is no redundancy of background information to improve robustness to foreground motion. %exist. %improve TRGMC robustness.
Therefore, we need a different means of outlier handling.
We handle this issue by assigning higher weights to links that are more likely to be connected to the background. %explicitly

For each keyframe $i$, we quantify how well the links emanating from frame $i$ are aligned with other keyframes.
If the alignment error is small, i.e., $\epsilon_k^{(i)} = \big|  \mathcal{W}_x(x_k^{(i)}, y_k^{(i)}; \p)  -    u_k^{(i)}   \big|   +    \big|  \mathcal{W}_y(x_k^{(i)}, y_k^{(i)}; \p)  -    v_k^{(i)}   \big|   <  \tau$, the link $k$ is more likely on the background of frame $i$ and thus, more reliable for aligning non-keyframes.
We create a \textit{reliability map} for each keyframe $i$, denoted as $\mathbf{R}^{(i)}$ (Fig.~\ref{fig:reliabilitymap}). %, of the same size as frame $i$ 
For each link $k$ with $\epsilon_k^{(i)} < \tau$, a Gaussian function with $\mu_k = (x_k^{(i)}, y_k^{(i)})$ and $\sigma_k=c s_k$ is superposed on $\mathbf{R}^{(i)}$, where the constant $c$ is $20$.
We define, % $\mathbf{R}^{(i)}_{m,n}$ as,
\eqnvspace
\begin{equation}
%\mathcal{R}^{i}(m,n)= \max \bigg( \eta , \min \Big(1,  \sum_{k \in \mathbb{B}_i}e^ {- \frac{\big( m-x_k^{(i)} \big)^2 + \big( n-y_k^{(i)} \big)^2}{2\sigma_k ^ 2}} \Big) \bigg),
\mathbf{R}^{(i)}_{m,n}= \Bigg \lceil  \bigg\lfloor \sum_{k \in \mathbb{B}_i}e^ {- \frac{\big( m-x_k^{(i)} \big)^2 + \big( n-y_k^{(i)} \big)^2}{2\sigma_k ^ 2}} \bigg\rfloor_1 \Bigg \rceil_\eta,
\eqnvspace
\end{equation}
where $\mathbb{B}_i = \{k | \epsilon_k^{(i)} < \tau\}$, $\eta > 0$ is a small constant (set to $0.1$), $\lceil x \rceil_{\eta}=\mbox{max}(x,\eta)$ and $\lfloor x \rfloor_1=\mbox{min}(1,\eta)$. % is the subset of keypoints with acceptable error and .
Now, we assign the weight of the links connecting frame $j$ to the keyframe $d_k^{(j)}$ at the coordinate $(u_k^{(j)} , v_k^{(j)})$, as the reliability map of the keyframe at the endpoint,
$\Omega^{(j)}_{k,k} = \big(\mathbf{R}^{(a)}_{u_k^{(j)}   ,   v_k^{(j)}} \big)^{r^q}$, where $a=d_k^{(j)}$.  
%\eqnvspace 
%\label{eqn:weight2}
%\end{equation}
%where $t$ is the iteration index, and $0<r<1$ controls rate of change of the weights. 

We summarize the TRGMC algorithm in Algorithm~\ref{alg:trgmc}.

\iffalse

\SubSection{TRGMC of long video sequences}
To align a long video sequence efficiently, we break down the input video to smaller temporal chunks and process them sequentially. 
Starting from the first chunk, TRGMC is applied to the chunk $c$ (as in Algorithm~\ref{alg:algorithm})  and then, the \textit{transparent} keypoints of chunk  $c$ are used to align with chunk $c+1$.
By transparent keypoints we mean the keypoints for which the pixel value at their spatio-temporal coordinate appears on the motion panorama.
In other words, at these keypoint locations, no future frame is overlaid on the motion panorama. 
The transformation found from applying TRGMC on chunk $c+1$ and the transparent keypoints of chunk $c$, is accumulated with the existing transformations of chunk $c$.
%Next, chunk $c+2$ is aligned with transparent keypoints of the chunks $c$ and $c+1$.
We repeat this procedure sequentially until all chunks are aligned relative to each other.

\fi
%{\color{red} Algorithm~\ref{alg:algorithm} summarizes the proposed TRGMC algorithm. WILL MODIFY THE ALGORITHM IF WE HAVE ENOUGH SPACE TO KEEP IT.}

\begin{algorithm} [t!]
\label{alg:algorithm}
\small
\KwData{A set of input frames $\{\mathbf{I}^{(m)}\}_{m=1}^M$}
\KwResult{A set of homography matrices $\{ \p_m \}_{m=1}^M$}
%Specify $N$ keyframes $\mathbb{K}=\{k_1, ..., k_N\}$ (sec.~\ref{sec:alignment})\;
%\vspace{2mm}
\tcc{Align keyframes (Sec.~\ref{sec:jointalignmentsolution})}
%Detect and match keypoints in consecutive keyframes\;
Specify $\mathbb{K}=\{k_1, ..., k_N\}$ and initialize (Sec.~\ref{sec:init})\; %the homographies 
Match keypoints of all frames $i \in \mathbb{K}$ densely\;
%Filter out matches outside approximate overlap region\;
Prune links (Sec.~\ref{sec:robustness}) and set weights (Eqn.~\ref{eqn:weights1})\;
Store links' start and end coordinates in $(\x_i,\y_i)$ and $(\mathbf{u}_i,\mathbf{v}_i)$\;		

\Repeat{$q < T_1$ \text{or} $ \big( \frac{1}{N}\sum_{i \in \mathbb{K}} ||\Delta \p_i||^2 > \tau_1 \big) $}
{
     	\ForAll{$i \in \mathbb{K}$}{
	Compute $\Delta \p_i$\ (Eqn.~\ref{eqn_solution}), update $\p_i$, $\x_i$ and $\y_i$ \;
%	$\p_i = \p_i + \Delta \p_i $\;
%	Update $\x_i$, $\y_i$ according to $\p_i$\;
%	\ForAll{$m \in \mathbb{K} \backslash \{i\}$}{
	Update $(\mathbf{u}_m$, $\mathbf{v}_m)$ according to $\p_i$ for $m \in \mathbb{K} \backslash \{i\}$\;		
	Update weights (Eqn.~\ref{eqn:weights1})\;
	}	
	{ $q \leftarrow q+1$\;}
%{\color{red}
%	\If{$\epsilon < \epsilon^{*}$} {
%		$\epsilon^{*} = \epsilon$, $\p_i^{*}=\p_i$ for all $i \in \mathbb{K}$\;
%	}
%}
 }
%\vspace{2mm}
\tcc{Align non-keyframes (Sec.~\ref{sec:nonkeyframes})}
Compute reliability map $\mathbf{R}^{(i)}$ for $ i \in \mathbb{K}$ \;
\For{$i=1:N-1$}{
	\ForAll{$j \in \{ k_{i}+1, ..., k_{i+1}-1  \}$}{
Match keypoints in $j$ with $d^{(j)} \in \{k_i,k_{i+1} \}$ \;
Prune links (Sec.~\ref{sec:robustness}) and set weights $\Omega^{(j)}_{k,k}$\; %(Eqn.~\ref{eqn:weight2})\;
Store links' coordinates in $(\x_j,\y_j)$ and $(\mathbf{u}_j,\mathbf{v}_j)$\;

%	Align frames between keyframes $k_{i}$ and $k_{i+1}$\;
	\Repeat{$q < T_2$ or $ \big( ||\Delta \p_j ||^2> \tau_2 \big) $}{
	Compute $\Delta \p_j$\ (Eqn.~\ref{eqn_solution}), update $\p_j$, $\x_j$ and $\y_j$\;
%	$\p_j = \p_j + \Delta \p_j $\;
%	Update $\x_j$, $\y_j$ according to $\p_j$\;
	Update weights (Eqn.~\ref{eqn:weights1}), $q \leftarrow q+1$\;  %(Eqn.~\ref{eqn:weight2})
	}

	}
}
\caption{\small TRGMC Algorithm}
\label{alg:trgmc}
\vspace{-1mm}
\end{algorithm}

%------------------------------------------------------------%
\Section{Experimental Results and Applications}
\label{sec:results}
We now present qualitative and quantitative results of the TRGMC algorithm and discuss how different computer vision applications will benefit from TRGMC.

\SubSection{Experiments and results}

\Paragraph{Baselines and details}
We choose three sequential GMC algorithms as the baselines for comparison: MLESAC~\cite{mle} and HEASK~\cite{heask} both based on our own implementation, and RGMC~\cite{rgmc} based on the authors' Matlab code available online.
We implement TRGMC in Matlab, and will publish the code.
Denoting the video frames of $w\times h$ pixels, we set the parameters as $\gamma = 0.1 w h$, $T_1=300$, $\tau_1=5 \times 10^{-4}$, $T_2=50$, $\tau_2=10^{-4}$, $r=0.7$, $\tau = 1$,
% $\Delta f_1=5$, $\Delta f_2=20$
$\Delta f=10$, and $\beta=1$.
For the backward-forward scheme we set $\alpha=1$ and for the backward scheme $\alpha=0$.

\Paragraph{Datasets and metric}
Given there is no public dataset for quantitative GMC evaluation, we form a dataset composed of $40$ challenging videos from SVW~\cite{svw} and $15$ videos from UCF101~\cite{ucf101}, termed ``quantitative dataset''.   %Sports Videos in the Wild (
SVW is an extremely unconstrained dataset including videos of amateurs practicing sports, and is also captured by amateurs via smartphone. %, and is very unconstrained.
%In addition, we use $15$ videos from UCF101 dataset~\cite{ucf101}. 
The min. and max. spatial size of videos are $240$ and $480$ pixels, respectively.
The average, min., and max. length of the videos are $14$, $3$, and $45$ seconds, captured at $25$ or $30$ FPS.
In addition, we form another ``qualitative dataset'' with $200$ \textit{unlabeled} videos from SVW, in challenging categories of boxing, diving, and hockey.

To compare GMC over different temporal distances of frames,  % FIXME   this is not the reason! we label 5 frames cause we do not hae time to label all frames. We do all pairs because composite transformation from the labeling of neighboring frames is not accurate!
for each video of length $M$ in the quantitative dataset, we manually align all $10$ possible pairs from the $5$-frame set, $\mathbb{F}=\{ 1, 0.25M, 0.5M, 0.75M, M\}$, as long as they are overlapping, and specify the background regions.
For this, a GUI is developed for a labeler to match $4$ points on each frame pair, and fine tune them up to a half-pixel accuracy, until the background difference is minimized. 
Then, the labeler selects the foreground regions which subsequently identify the background region.
Similar to~\cite{rgmc}, we quantify the consistency of two warped frames $\I^{(i)}(\p_i)$ and $\I^{(j)}(\p_j)$ ($0$ to $1$ grayscale pixels) via the background region error (BRE), 
\eqnvspace
%\small
\begin{equation}
\mbox{BRE}(i,j) = \frac{1}{\Arrowvert\mathbf{M_B}\Arrowvert_1}{\Arrowvert | (\I^{(i)}(\p_i)-\I^{(j)}(\p_j)) |\odot\mathbf{M_B}  \Arrowvert_1},
%\mbox{BRE}(i,j) = \frac{\Arrowvert | (\I^{(i)}(\p_i)-\I^{(j)}(\p_j)) |\odot\mathbf{M_B}  \Arrowvert_1}{\Arrowvert\mathbf{M_B}\Arrowvert_1},
\eqnvspace
\end{equation}%\normalsize
where $\odot$ is element-wise multiplication and $\mathbf{M_B}$ is the background mask for the intersection of two warped frames.

\begin{table} [t!]
\begin{center}
\footnotesize
\setlength\tabcolsep{3.8pt}
\begin{tabular} {|c|c|c|c|cc|c|}
\hline
Algorithm & MLESAC & HEASK & RGMC & \multicolumn{2}{c|}{TRGMC} & GT*\\ \hline
Setting & -- & -- & -- & BF* & B* & --  \\ \hline
Avg. BRE  & 0.116 & 0.110 & 0.097 & \textbf{0.058} & 0.060 & 0.038  \\ \hline
Efficiency (s/f)   & 0.17 & 7.47 & 3.47 & 0.64 & 0.41 & -- \\ \hline
\end{tabular}
\vspace{1mm}
\caption{\small Comparison of GMC algorithms on quantitative dataset (*GT: Ground truth, BF: Backward-Forward, B: Backward).}
\figvspace\figvspace
\label{table:error} 
\end{center}
\end{table}

\begin{figure} [t!]
\centering
\includegraphics[trim=2cm 0cm 2cm 0cm,clip, width=.7\columnwidth]{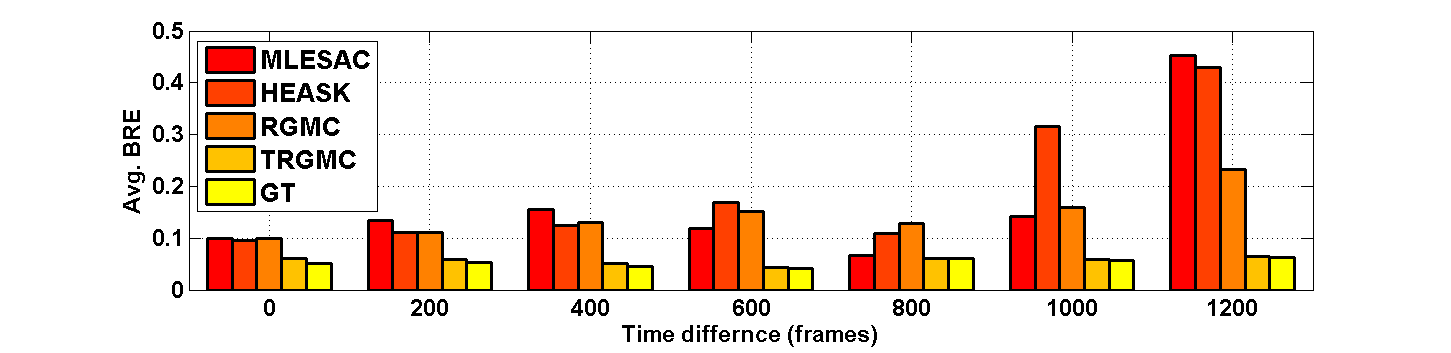} 
\vspace{-4mm}
\caption{\small Average BRE of frame pairs versus the time difference between the two frames.}
\label{fig:temporalError}\vspace{-3mm}
\end{figure}

\begin{figure*} [t!]
\begin{center}
%\begin{tabular}{p{0.3cm}p{2.65cm}p{0.3cm}p{3cm}p{0.3cm}p{3cm}p{0.3cm}p{2.5cm}}
\begin{tabular}{p{0.28cm}p{2.4cm}p{0.28cm}p{2.3cm}p{0.3cm}p{2.6cm}p{0.32cm}p{2.5cm}}
%\vspace{-2mm}
%\begin{tabular}{llllllll}
\small(a)&
\includegraphics[trim=0cm .1cm 0cm -1cm,clip,height=1.8cm]{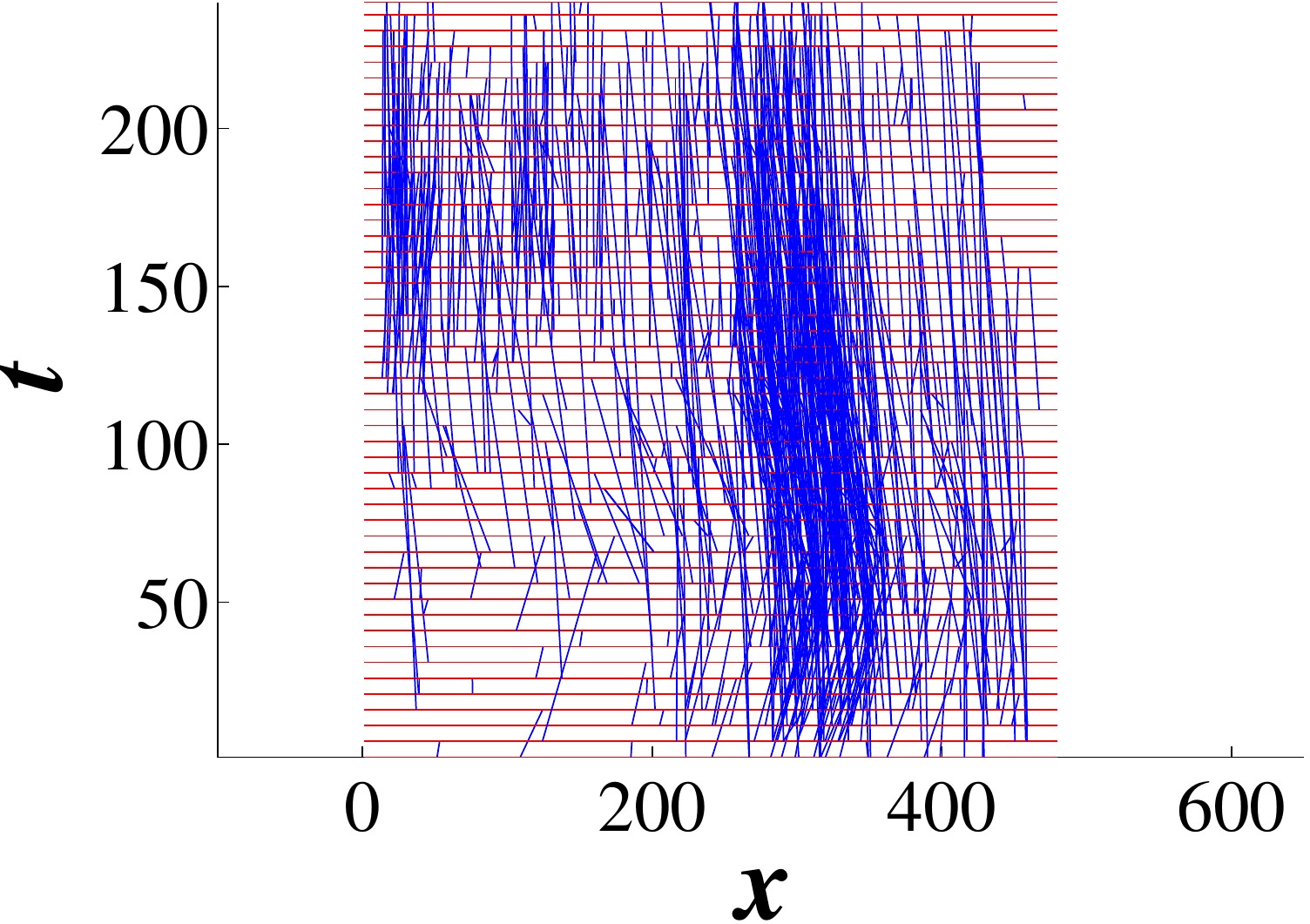} 
&
\small(b)&
\includegraphics[trim=0cm .1cm 0cm -1cm,clip,height=1.8cm]{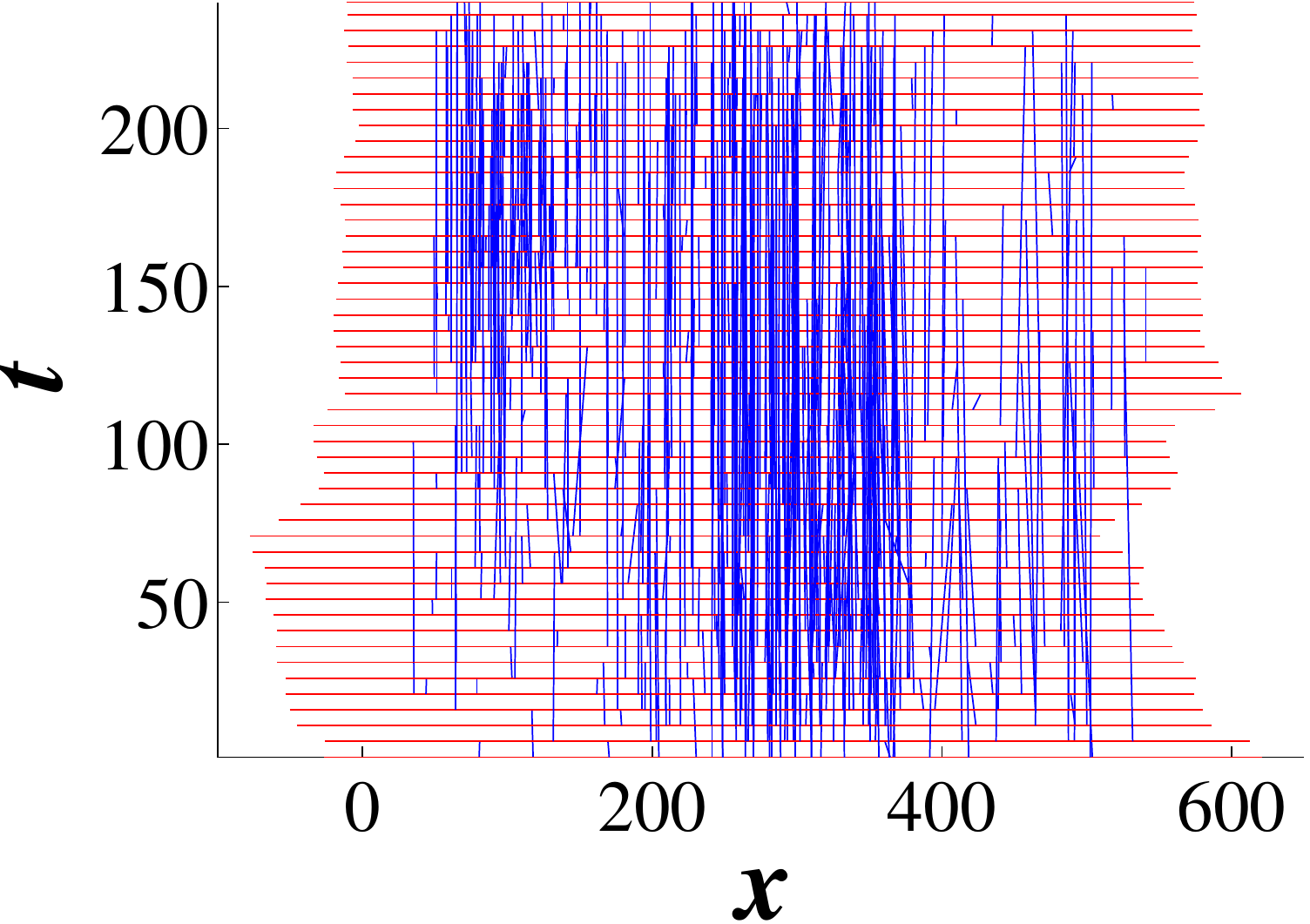}
&
\small(c)&
\includegraphics[trim=0cm 0cm 1.6cm 1.6cm,clip,height=1.65cm]{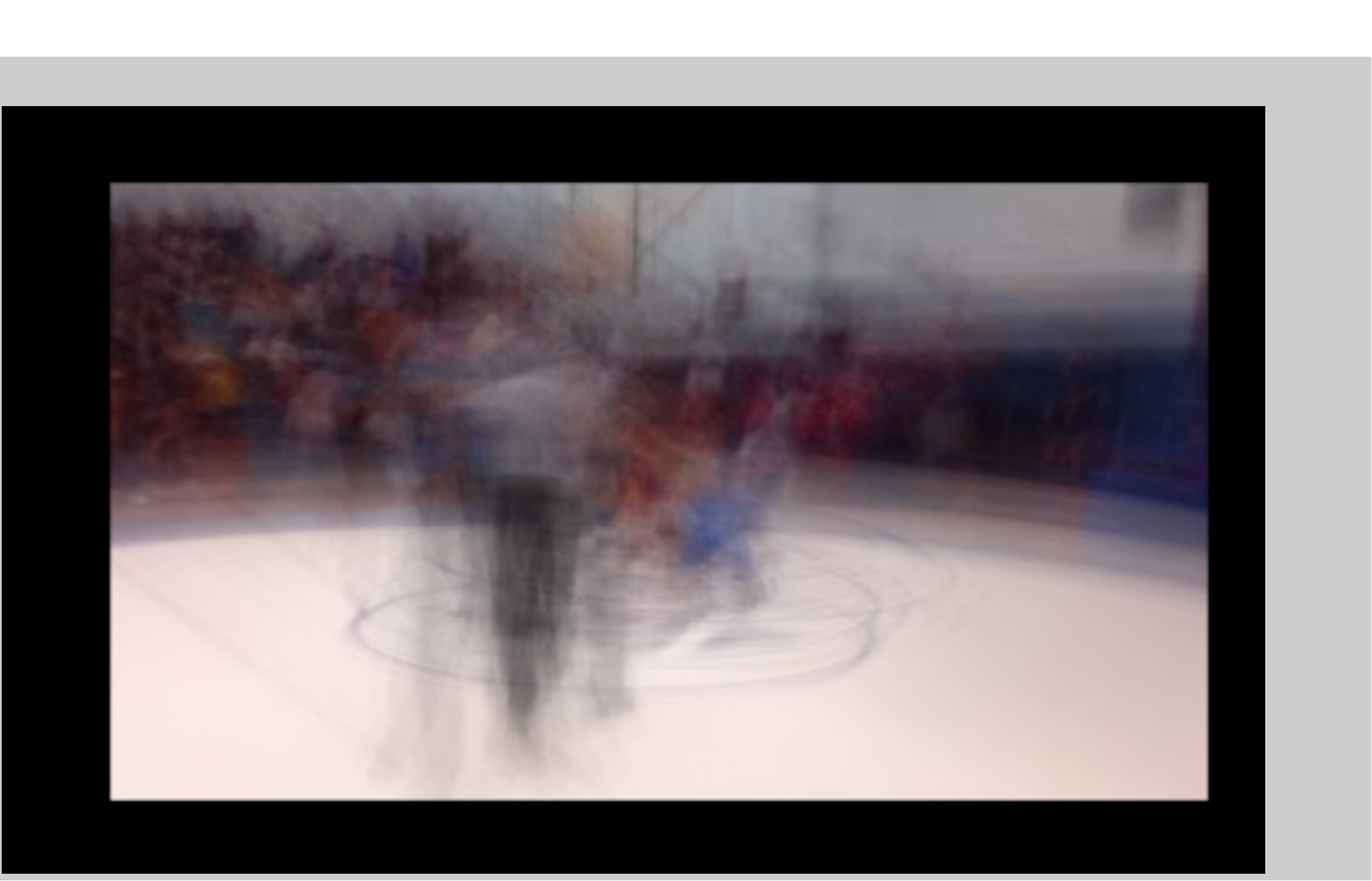} 
&
\small(d)&
\includegraphics[trim=2.8cm 1.5cm 2.5cm 1cm,clip,height=1.65cm]{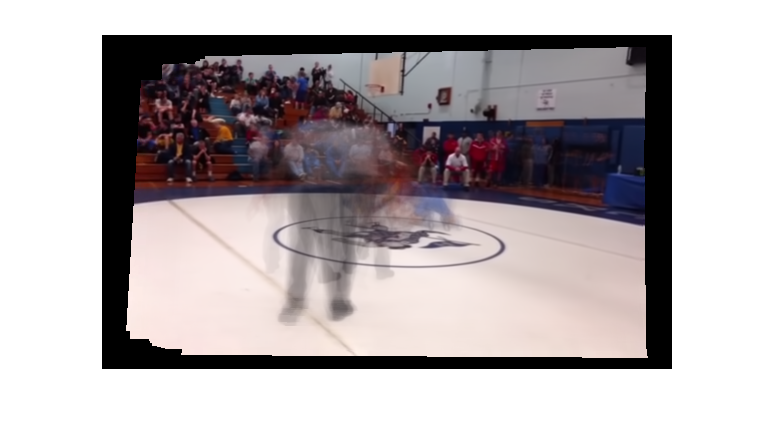} 
%\small         (a) & \small (b)  & \small (c)  & \small (d) 

\end{tabular}
\end{center}
\vspace{-10mm}
\caption{\small{Top view of the frames and links (a) before and (b) after TRGMC. The parallel links in (b) show successful \textit{spatial} alignment of keypoints. Average of frames (c) before and (d) after TRGMC. For better visibility, we show up to $15$ links emanated per frame.}} % To better illustrates performance, initialization module is disabled. Frames are warped after application of TRGMC, so length of frames in the top view is different from the input frames.)}}
\label{fig:interconnection}
%\figvspace
\vspace{-5mm}
\end{figure*}

\begin{figure*} [t]
\centering
\begin{tabular}{clclclcl}

%\multicolumn{3}{}{}   \includegraphics[trim=.5cm 1.2cm 4.5cm 2.2cm,clip,height=2.7cm]{images/26185_2_startend.pdf}
  
%  &

\multicolumn{2}{}{}   \includegraphics[trim=.3cm 1cm 2cm 2.4cm,clip,height=1.63cm]{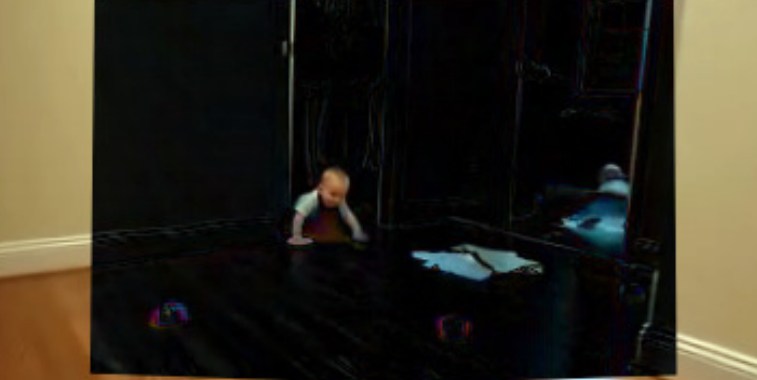}

 &

\multicolumn{2}{}{}   \includegraphics[trim=0cm 1.8cm 9.6cm 2.5cm,clip,height=1.63cm]{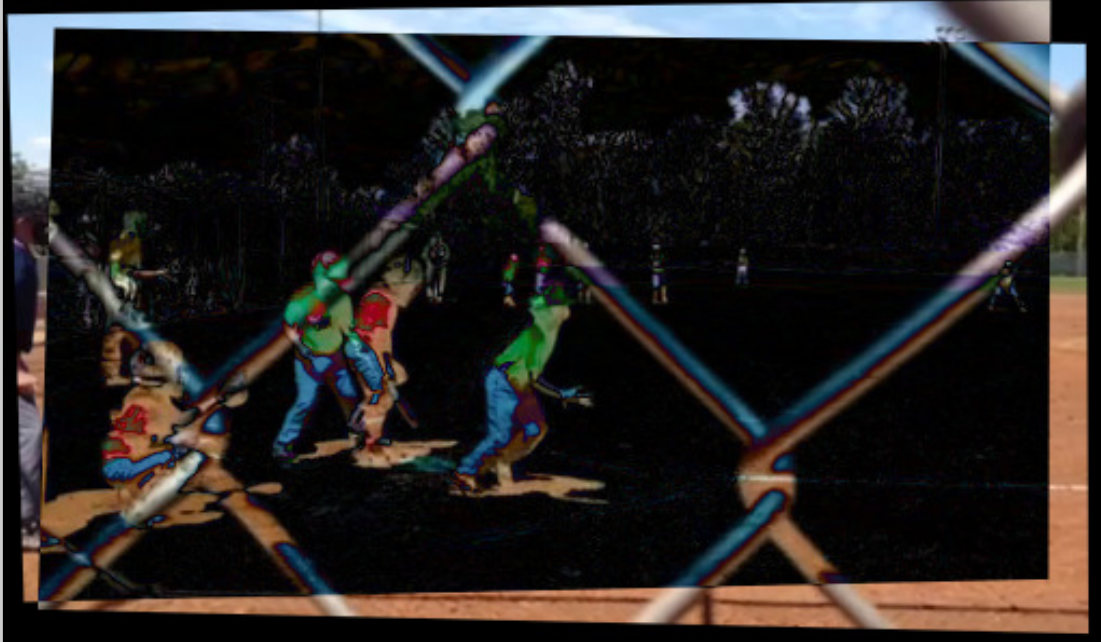}

&

\multicolumn{2}{}{}   \includegraphics[trim=3.3cm 1.5cm 4.5cm 2cm,clip,height=1.63cm]{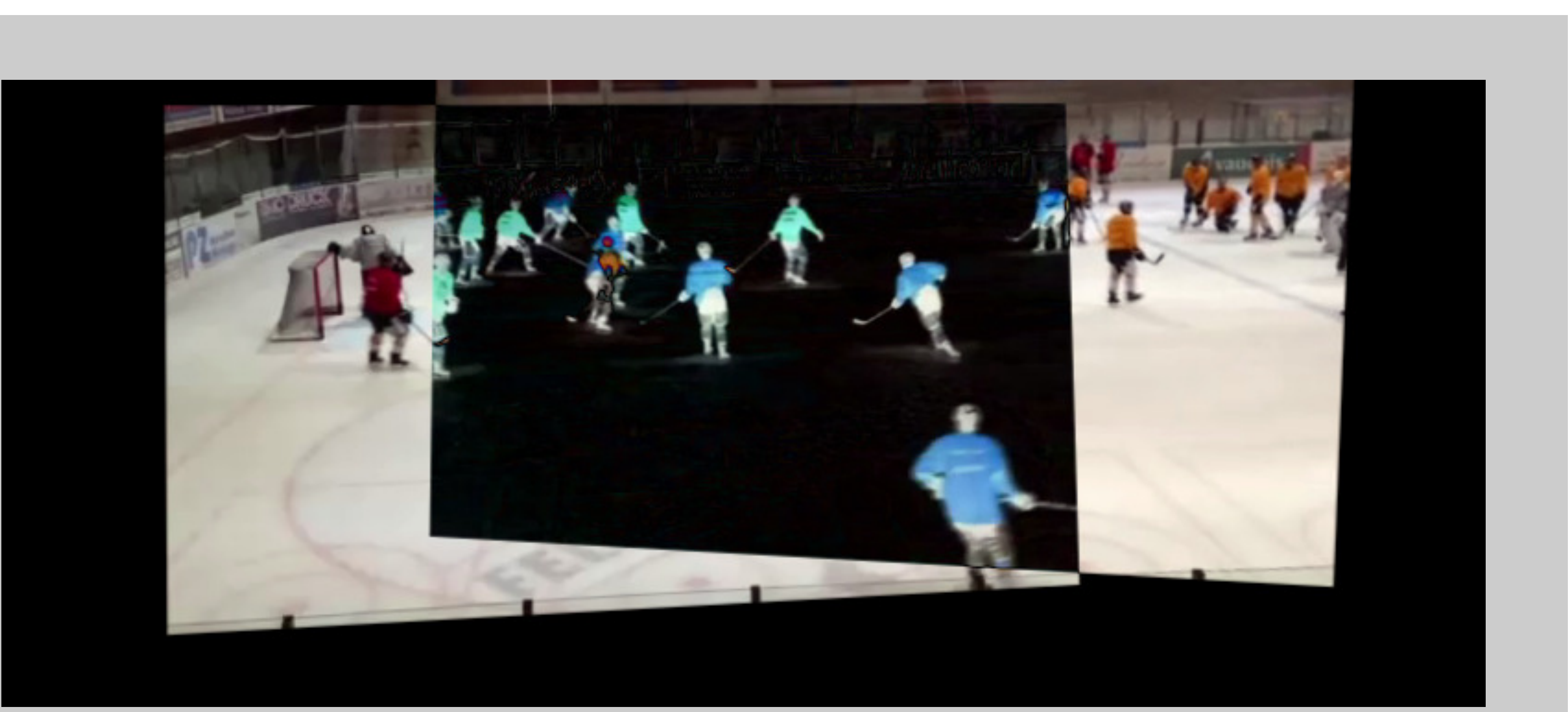}

&

\multicolumn{1}{}{}   \includegraphics[trim=0cm 0cm 1.6cm 1.7cm,clip,height=1.63cm]{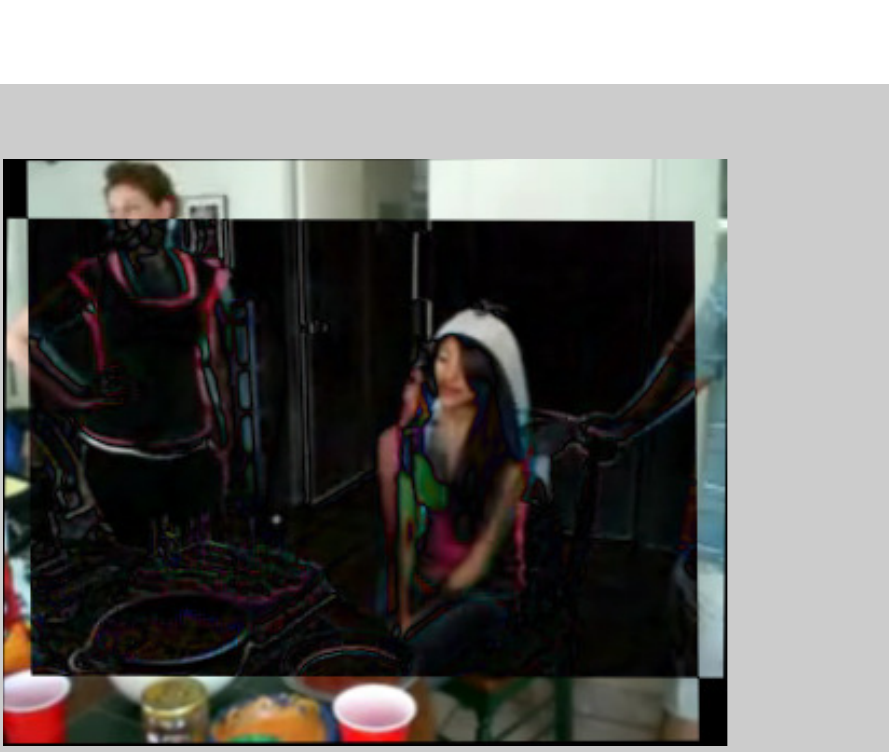}

%\multicolumn{2}{}{}   \includegraphics[trim=3.3cm 1.5cm 10cm 2cm,clip,height=2.41cm]{images/986_startend.pdf}
%\multicolumn{2}{}{}   \includegraphics[trim=1cm 0cm 4.5cm 1.8cm,clip,height=2.41cm]{images/4529_startend.pdf}
\\

\end{tabular}
\vspace{-4mm}
\caption{\small Composite image formed by overlaying the frame $n$ on frame $1$ for several videos after TRGMC. Left to right, $n$ is equal to $144$, $489$, $912$, $93$, respectively. In the overlap region the difference between the frames is shown.}
\label{fig:framedifference}
\figvspace
\vspace{1mm}
\end{figure*}

\Paragraph{Quantitative evaluation}
Average of BRE over all the temporal frames pairs is shown in Table~\ref{table:error}. 
%It is worth noting that for all the algorithms, we have reported the average performance for the videos with successful termination of the algorithm.
%For MLESAC, HEASK, RGMC, and TRGMC there has been $10$, $1$, $4$, and $2$ videos respectively, for which the code has crashed.
%This is mostly due to excessive memory consumption when the estimation is extensively incorrect, resulting the extremely large size of the warped frames.
TRGMC outperform all the baseline methods with considerable margin.
%For TRGMC, we report results for~\textit{backward-forward (BF)} and~\textit{backward (B)} scheme. 
The~\textit{backward-forward (BF)} scheme has a slightly better accuracy than the~\textit{backward (B)} scheme, and is also more stable based on our visual observation.
Thus, we use BF as the default scheme for TRGMC.

To illustrate how the accumulation of errors over time affects the final error, Fig.~\ref{fig:temporalError} summarizes the average error versus the time difference between the frames in $\mathbb{F}$.
This shows that TRGMC error is almost constant over a wide temporal distance between the frames.
Thus, even if a frame is not aligned accurately, the error is not propagating to all the frames after that.
However, in sequential GMC, the error increases as the time difference increases.

\iffalse
\begin{table} [t!]
\begin{center}
\footnotesize
\setlength\tabcolsep{3.8pt}
\begin{tabular} {|c|c|c|c|}
\hline

%\diaghead{\theadfont Performance}%
%{Algorithm}{Performance}&
%\thead{Second\\Good}&\thead{Third\\Shaky}&\thead{Fourth\\Failed}\

Algorithm \textbackslash  Perfromance & Good & Shaking& Failed \\ \hline
RGMC & 64\% & 33\% & 3\% \\ \hline
TRGMC  & \textbf{93\%}& 5\% & 2\% \\ \hline
\end{tabular}
\vspace{1mm}
\caption{\small Comparison of GMC algorithms on qualitative dataset.}  
\figvspace
\figvspace
\label{table:errorqualitative} 
\end{center}
\end{table}
\fi

\Paragraph{Qualitative evaluation}
While quantitative results are comprehensive, the number of videos is limited by the labeling cost.
Thus, we further compare TRGMC and the best performing baseline, RGMC, on the larger qualitative dataset.
The resultant motion panoramas were \textit{visually} investigated and  categorized into three cases: good, shaking, and failed (i.e., considerable background discontinuity). 
The comparison in Tab.~\ref{table:errorqualitative} again shows the superiority of TRGMC.

Figure~\ref{fig:interconnection} shows the \textit{links} of a sample video processed by TRGMC, and the average frames, before and after processing. 
Initialization module is disable for generating this figure to better illustrate how well the spatial coordinate of the keypoints are aligned, resulting in links parallel to the $t-$ axis.
This video also shows how GMC might be utilized for video stabilization.
Figure~\ref{fig:framedifference} shows a composite image formed by overlaying the last frame (or a far apart frame with enough overlap) on frame $1$ for several videos, after TRGMC. 
In the overlap region, difference between the two frames is shown, to demonstrate how well the background region matches for the frames with large temporal distance.
%Also, we apply RGMC and TRGMC on $200$ \textit{unlabeled} videos from SVW dataset (in challenging categories of boxing, diving, and hockey).
%The resultant motion panoramas were \textit{visually} investigated, and failure rates of $2\%$ and $3\%$ were observed for RGMC and TRGMC, respectively.
%In this study, considerable motion on the background region has been interpreted as failure.
%On the other hand, for RGMC and TRGMC, $5\%$ and $35\%$ of videos suffer from shaking.

\Paragraph{Computational efficiency}
Table~\ref{table:error} also presents the average time for processing each frame for each method, on a PC with an Intel i$5$-$3470$@$3.2$GHz CPU, and $8$GB RAM.
While obtaining considerably better accuracy than HEASK or RGMC, TRGMC is on average $15$ times faster than HEASK and $7$ times faster than RGMC.
MLESAC is $\sim$$3$ times faster than TRGMC, but with twice the error.
For TRGMC, the backward scheme is $50\%$ faster than forward-backward, since it has approximately half the links of BF.

\begin{figure} [t!]
\centering
%\includegraphics[trim=0cm 0cm 0cm 0cm,clip,height=2cm]{images/steps.pdf} 
%\vspace{-2mm}
%\caption{\small Error and efficiency vs.~the keyframe selection step, $\Delta f$.}
%\label{fig:tradeoff}

\begin{minipage}[t]{0.4\linewidth}
    \includegraphics[trim=0cm 0cm 0cm 0cm,clip,height=1.8cm]{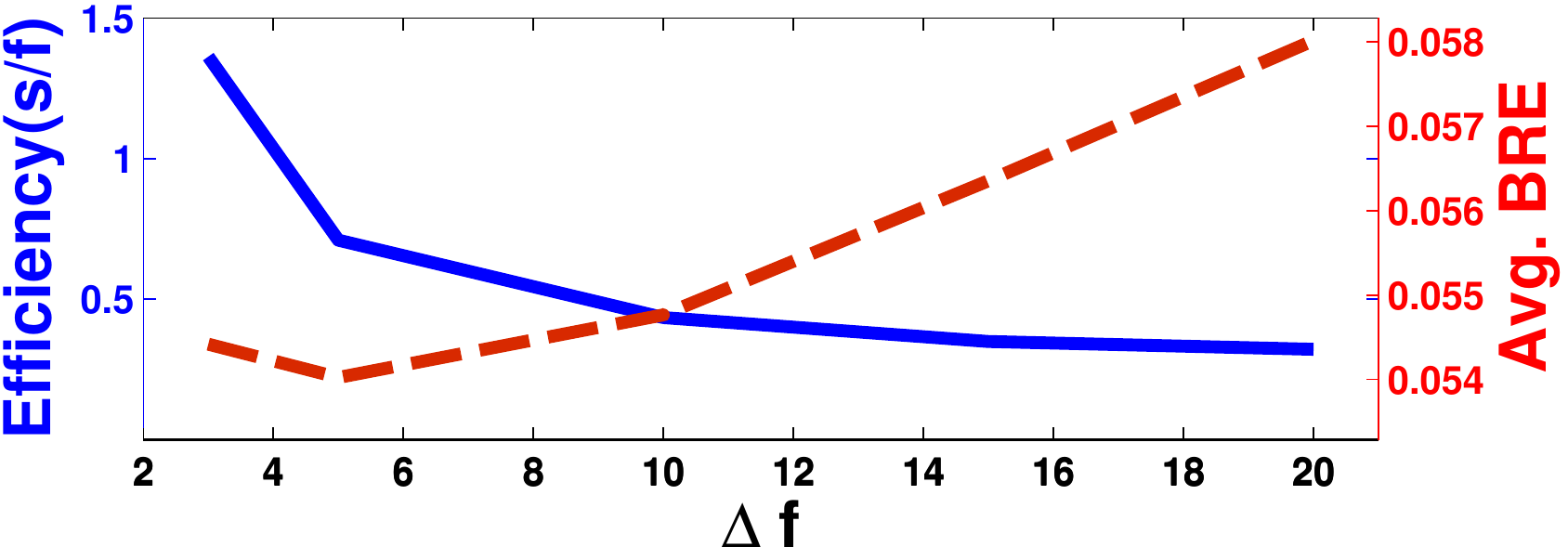} 
    \vspace{-8mm}
	\caption{\small Error and efficiency vs.~the keyframe selection step, $\Delta f$.}
	\label{fig:tradeoff}
\end{minipage}%
    \hfill%
\begin{minipage}[t]{0.5\linewidth}
\footnotesize
\setlength\tabcolsep{1.8pt}
\vspace{-16mm}
\begin{center}
\begin{tabular} {|c|c|c|c|}
\hline
\scriptsize
%\diaghead{\theadfont Performance}%
%{Algorithm}{Performance}&
%\thead{Second\\Good}&\thead{Third\\Shaky}&\thead{Fourth\\Failed}\

Alg. \textbackslash  Perfromance & Good & Shaking& Failed \\ \hline
RGMC & 64\% & 33\% & 3\% \\ \hline
TRGMC  & \textbf{93\%}& 5\% & 2\% \\ \hline
\end{tabular}
\vspace{-2.2mm}
\end{center}
\captionof{table}{\small Comparison of GMC algorithms on qualitative dataset.}  
\figvspace
\figvspace
\label{table:errorqualitative} 
\end{minipage} 
\figvspace
\end{figure}

\Paragraph{Accuracy vs.~efficiency trade-off}
Fig.~\ref{fig:tradeoff} presents the error and efficiency results for a set of $5$ videos versus the keyframe selection step, $\Delta f$.
For this set, the ground truth error is $0.049$.
As a sweet spot in the error and efficiency trade-off, we use $\Delta f=10$ for TRGMC.
This figure also justifies the two stage processing scheme in TRGMC, as processing frames at a low selection step $\Delta f$, is costly in terms of efficiency, but only improves the accuracy slightly.

\setlength{\fboxsep}{0pt}
\setlength{\fboxrule}{1pt}

\begin{figure*} [t!]
\centering \vspace{3mm}
\begin{tabular}{lllllllllll}

\multicolumn{3}{}{}   
  \begin{overpic}[trim=.5cm .5cm 8cm 2.3cm,clip,height=1.5cm]{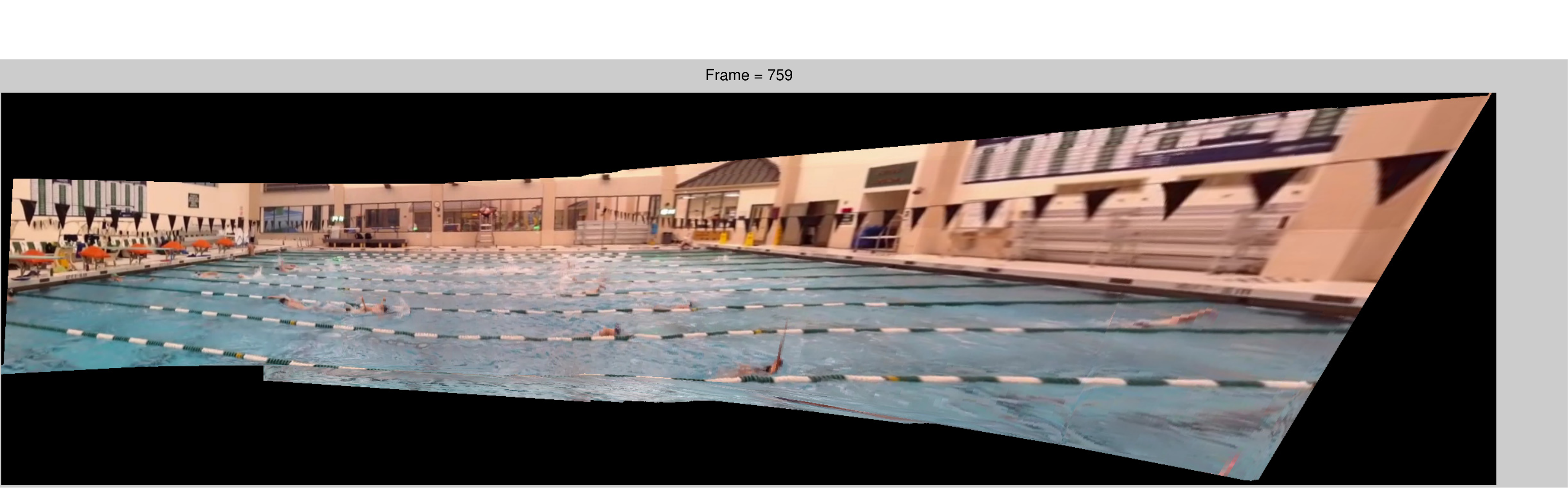}
     \put(2,2){\cfbox{green}{\includegraphics[trim=2cm 1cm 2cm 2cm,clip,height=.4cm]{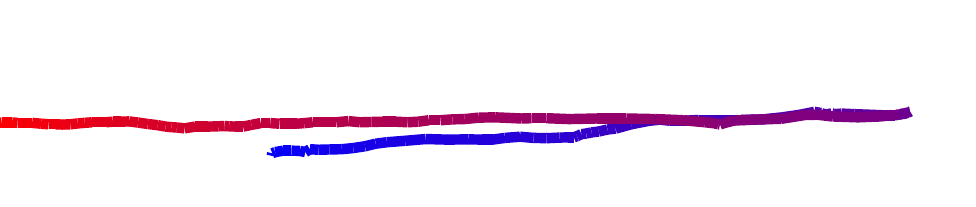}}}  
  \end{overpic}   
  
%\multicolumn{1}{}{} 
%  \begin{overpic}[trim=.3cm 0cm 2cm 2.2cm,clip,height=2.35cm]{images/4259.pdf}
%     \put(63,2){\cfbox{green}{\includegraphics[trim=0cm 0cm 1.5cm 1.5cm,clip,height=1cm]{images/4259_border.pdf}}}  
%  \end{overpic} 

  &
\multicolumn{1}{}{}   
  \begin{overpic}[trim=.5cm 3cm 1.5cm 2cm,clip,height=1.5cm]{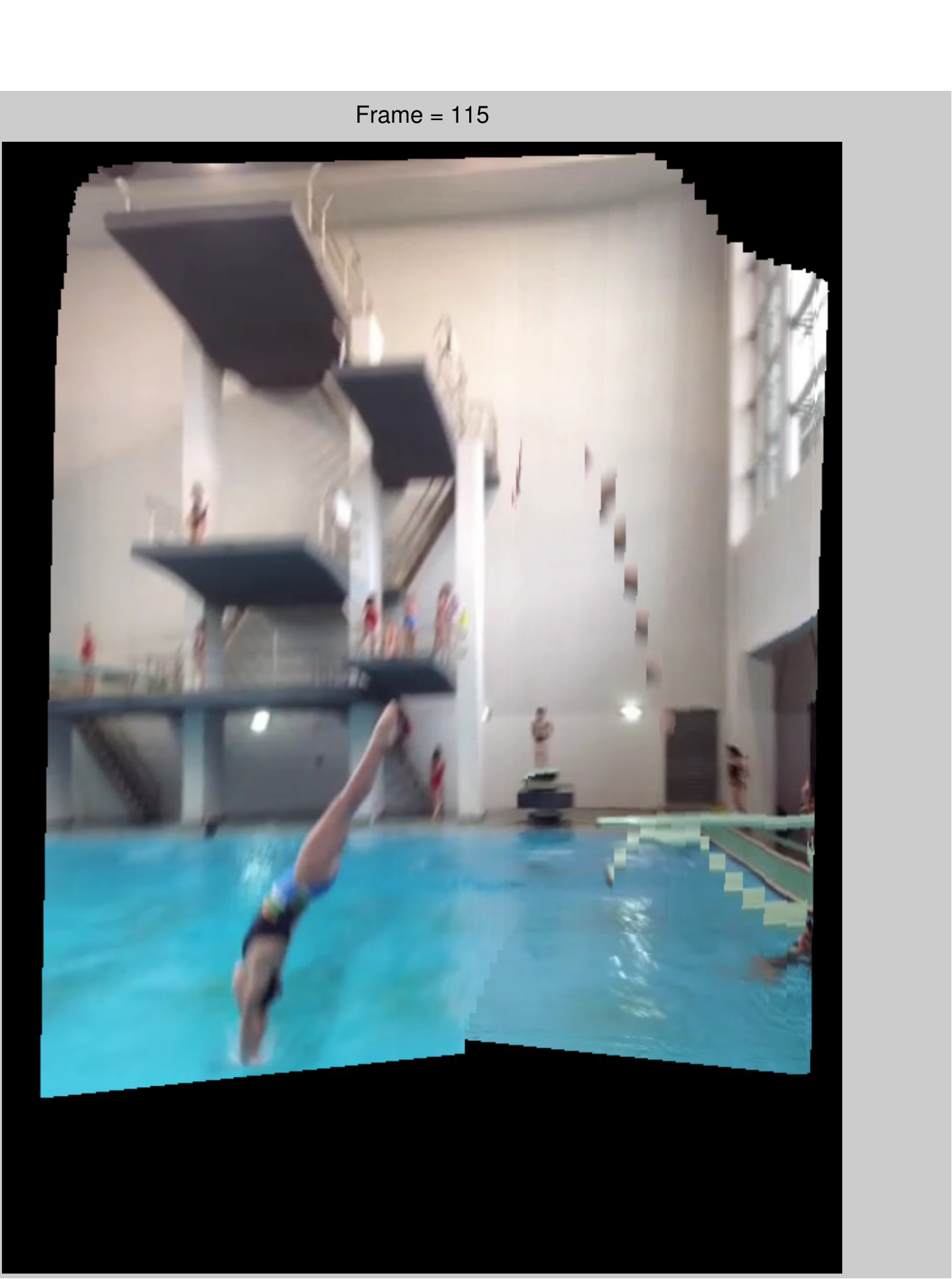}
     \put(37,2){\cfbox{green}{\includegraphics[trim=2cm 4cm 4cm 4cm,clip,height=.4cm]{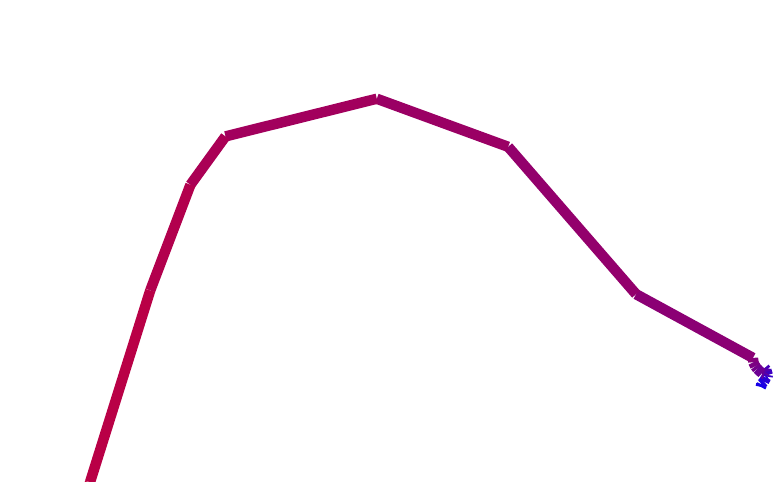}}}  
  \end{overpic} 

%\multicolumn{4}{}{} 
%  \begin{overpic}[trim=0cm 0cm 2cm 2.2cm,clip,height=2.35cm]{images/2285.pdf}
%     \put(56,2){\cfbox{green}{\includegraphics[trim=0cm 0cm 2cm 2cm,clip,height=1cm]{images/2285_border.pdf}}}  
%  \end{overpic} 

%\multicolumn{2}{}{} 
%  \begin{overpic}[trim=0cm 0cm 2cm 2cm,clip,height=2.23cm]{images/24203.pdf}
%     \put(4,45){\cfbox{green}{\includegraphics[trim=2cm 4cm 2cm 4cm,clip,height=.5cm]{images/24203_center.pdf}}}  
%  \end{overpic} 

%\multicolumn{2}{}{} 
%  \begin{overpic}[trim=0cm 2.5cm 2cm 2.2cm,clip,height=2.23cm]{images/2827.pdf}
%     \put(3,3){\cfbox{green}{\includegraphics[trim=2cm 2cm 2cm 2cm,clip,height=1cm]{images/2827_center.pdf}}}  
%  \end{overpic} 

%\multicolumn{4}{}{} 
%  \begin{overpic}[trim=1cm 0cm 3cm 2.2cm,clip,height=2.35cm]{images/00000.pdf}
%     \put(40,2){\cfbox{green}{\includegraphics[trim=2cm 4cm 2cm 4cm,clip,height=.5cm]{images/00000_center.pdf}}}  
%  \end{overpic} 
&  
\multicolumn{3}{}{}   
  \begin{overpic}[trim=1cm 1.1cm 3cm 2.2cm,clip,height=1.5cm]{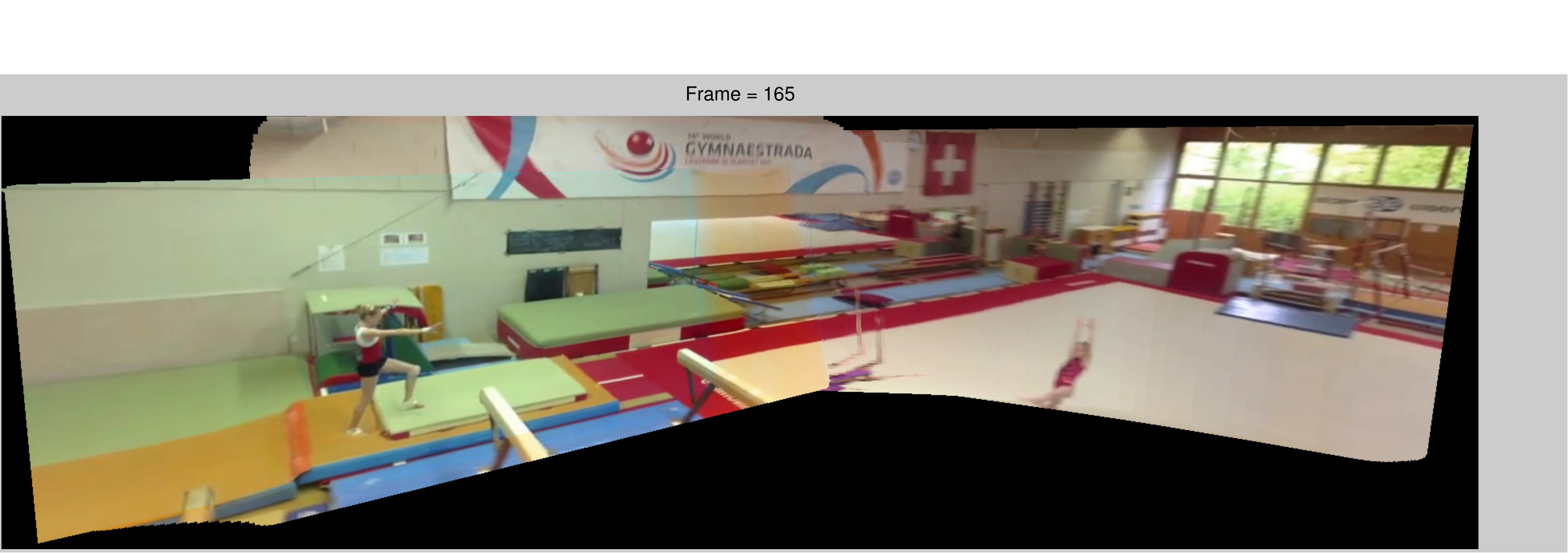}
     \put(50,1){\cfbox{green}{\includegraphics[trim=2cm 4cm 2cm 4cm,clip,height=.5cm]{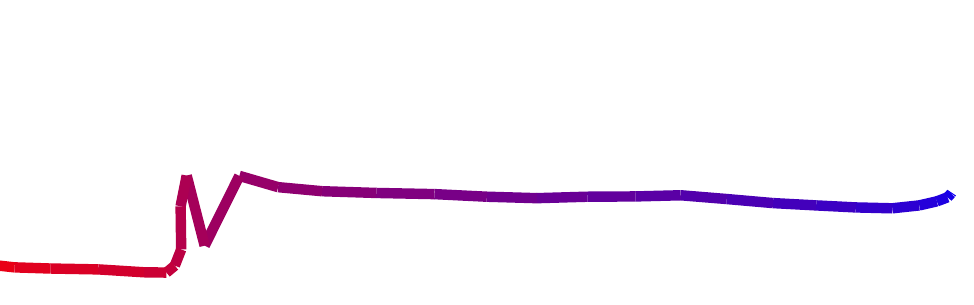}}}  
  \end{overpic}   
  \\  

\end{tabular}
\vspace{-3mm}
\caption{\small Temporal overlay of frames from different videos processed by TRGMC. Trajectory of the center of image plane over time is overlaid on each plot to show the camera motion pattern, where color changes from blue to red with progression of time.}
\label{fig:result}
\figvspace\vspace{1mm}
\end{figure*}

\SubSection{TRGMC applications}
\label{subsection:applications}

\Paragraph{Motion panorama}
By sequentially reading input frames, applying the transformation found by TRGMC, and overlaying the warped frames on a sufficiently large canvas, a motion panorama is generated. 
Furthermore, it is possible to reconstruct the background using the warped frames~\textit{first} (as will be discussed later), and overlay the frames on that, to create a more impressive panorama. 
The last frame on the video generated such, can be referred to as a panoramic mosaic~\cite{steedly2005efficiently}. 
Figure~\ref{fig:result} shows a few exemplar panoramas along with the camera motion pattern.
For all the input videos of length $M$, we apply $(\frac{1}{2}(\p_1 + \p_M))^{-1}$ to the transformations found by TRGMC to normalize the result and have a better view of the scene in a smaller spatial area.%, especially in presence of considerable camera panning.
% FIXME, "a better view " is not clear. Can we say making the middle frame in the center of the canvas?

\Paragraph{Raster scan of scenes/Image mosaic}
We may swipe the camera through a large scene in a raster scan fashion and use TRGMC to reconstruct a big image mosaic.
Note that this scenario is non-trivial since the accumulated error can be obvious when the raster scan comes back to the original camera position. 
The long term robustness presented by TRGMC is  crucial in this scenario.
%The same method used in generation of a motion panorama is used for this problem.
%Having many frames with considerable overlap stitched using TRGMC prevents many issues present in IS for low-overlap images.
%Some methods for motion deblurring such as~\cite{li2010generating} may be used for better quality.
%The long term robustness presented by TRGMC is crucial in this scenario.
%Usefulness of raster scan may be either due to high resolution requirement or small distance of camera from the scene (e.g. capturing image from a white board, or an artwork on the ground while walking). 

\begin{figure} [t!]
%\centering
\begin{tabular}{p{6.2cm}p{2.6cm}p{3.6cm}}
%\begin{tabular}{llll}

%\includegraphics[trim=2.7cm 1.5cm 2.6cm 1cm,clip,width=3.6cm]{images/bg_noWeight2.png} 
 
%& \small(a) & \small(b)
%\includegraphics[trim=3.2cm 2cm 2.6cm 1.5cm,clip,width=4cm]{images/bg_noWeight3.png}
%&
%\includegraphics[trim=3.2cm 2cm 2.6cm 1.5cm,clip,width=4cm]{images/bg_reliabilityWeighted3.png} 
%\\

%\small(c) & \multicolumn{2}{c}{
%\includegraphics[trim=2.7cm 2.5cm 10cm 2cm,clip,width=7.55cm]{images/bg_noWeight4.png}
%}
%\\
%\small(d) & \multicolumn{2}{c}{

%\multicolumn{1}{l}{
%\small(a)
%}

%\multicolumn{3}{l}{
\includegraphics[trim=3cm 2.5cm 10cm 2cm,clip,height=1.65cm]{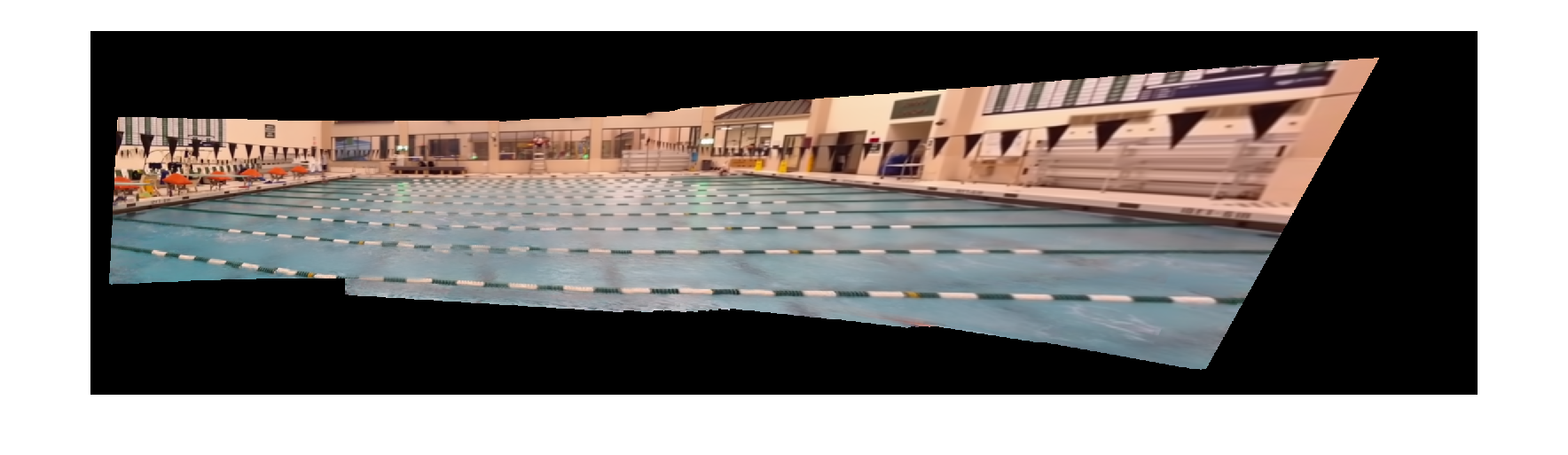} 
%}
%\\
%\small(b)
&
%\multicolumn{1}{c}{
\includegraphics[trim=2.7cm 1.5cm 2.6cm 1cm,clip,height=1.65cm]{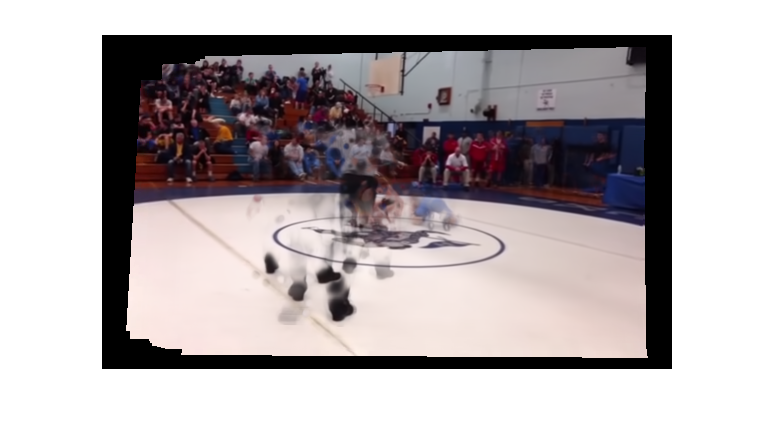} 
%}
%&
%\small(c)
&
\includegraphics[trim=1cm 1cm 2.1cm 2.5cm,clip,height=1.65cm]{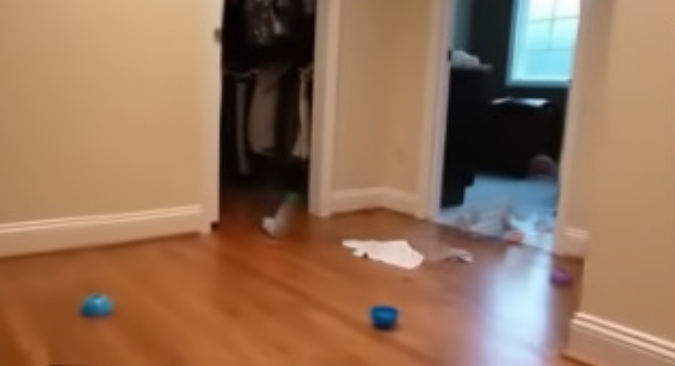} 
\\
\end{tabular}
\vspace{-4mm}
\caption{\small Background reconstruction results. Compare the left image with Fig.~\ref{fig:result}, middle image with Fig.~\ref{fig:interconnection}, and right image with Fig.~\ref{fig:framedifference}.}
\label{fig:bgreconstruct}
\vspace{-4mm}
\end{figure}

\Paragraph{Background reconstruction} 
Background reconstruction is important for removing occlusions, or detecting foreground~\cite{monari2011real}.  % by background subtraction
%To reconstruct the background, it is necessary to identify background pixels.
%A commonly used method for static cameras is to use a temporal median-filter over all the input frames.
%Thus, if a background region is visible for more than half of the frames, that region is correctly constructed. 
%In presences of foreground, the median shifts toward the foreground, thus \textit{mode} value might be used.%, i.e. the most frequently observed color, might be used.
%However, both methods require considerable memory and also rely on the background being visible in majority of the frames.
To reconstruct the background, a weighted average scheme is used to weight each frame by the \textit{reliability map}, $\mathbf{R}^{(i)}$,  which assigns higher weights to background.
Since the minimum value of $\mathbf{R}^{(i)}$ is a positive constant $\eta$, if no reliable keyframe exists at a coordinate, all the frames will have equal weights. 
Specifically, the background is reconstructed by
%\begin{equation}
$\mathbf{B} = \frac{\sum_{i \in \mathbb{K}} \mathbf{R}^{(i)}(\p_i) \mathbf{I}^{(i)}(\p_i)}
{\sum_{i \in \mathbb{K}} \mathbf{R}^{(i)}(\p_i)}$,
%\end{equation}    
where $\mathbf{R}^{(i)}(\p_i)$ and $\mathbf{I}^i(\p_i)$ are the reliability map and the input frame warped using the transformation $\p_i$.
%Fig.~\ref{fig:bgreconstruct} shows the reconstruction results. % and comparison with the average of the frames.  
Using our scheme, reconstructed background in Fig.~\ref{fig:bgreconstruct} is sharper and less impacted by the foreground.
%Given the reconstructed background, it is possible to render each warped frame on the background, resulting to a video with an impressive wide view of the scene spanned by the camera motion.

% FIXME: move (a,b) to be beside images,   also, there is no much difference between c and d.

\begin{figure} [t!]
\centering
\begin{tabular}{p{0.4cm}p{3.5cm}p{0.4cm}p{3.5cm}p{0.4cm}p{3.5cm}}
\small(a) & 
\includegraphics[trim=0cm 0cm 1.8cm 2cm,clip,width=3.4cm]{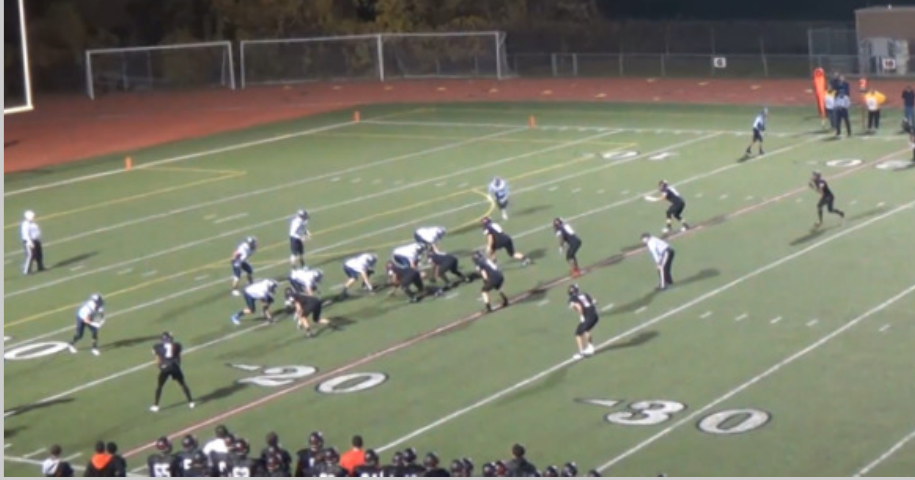} &
\small(b) & 
\includegraphics[trim=0cm 0cm 1.8cm 2cm,clip,width=3.4cm]{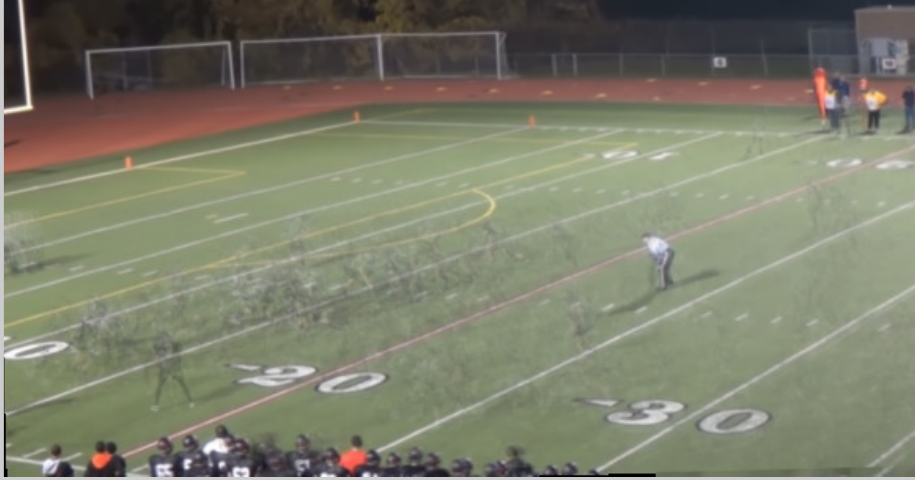} &
%\ \ \ \ \ \ \ \ \ \ \ \ \ \ \ \ \ \ \ \small(a) & 
%\ \ \ \ \ \ \ \ \ \ \ \ \ \ \ \ \ \ \ \ \ \small(b) \\
%%\small(c) & 
%%\includegraphics[trim=.7cm .65cm 1.7cm 2cm,clip,height=2cm,width=3.51cm]{images/fg_diff.pdf} &
\small(c) & 
\includegraphics[trim=0cm 0cm 1.72cm 2cm,clip,width=3.4cm]{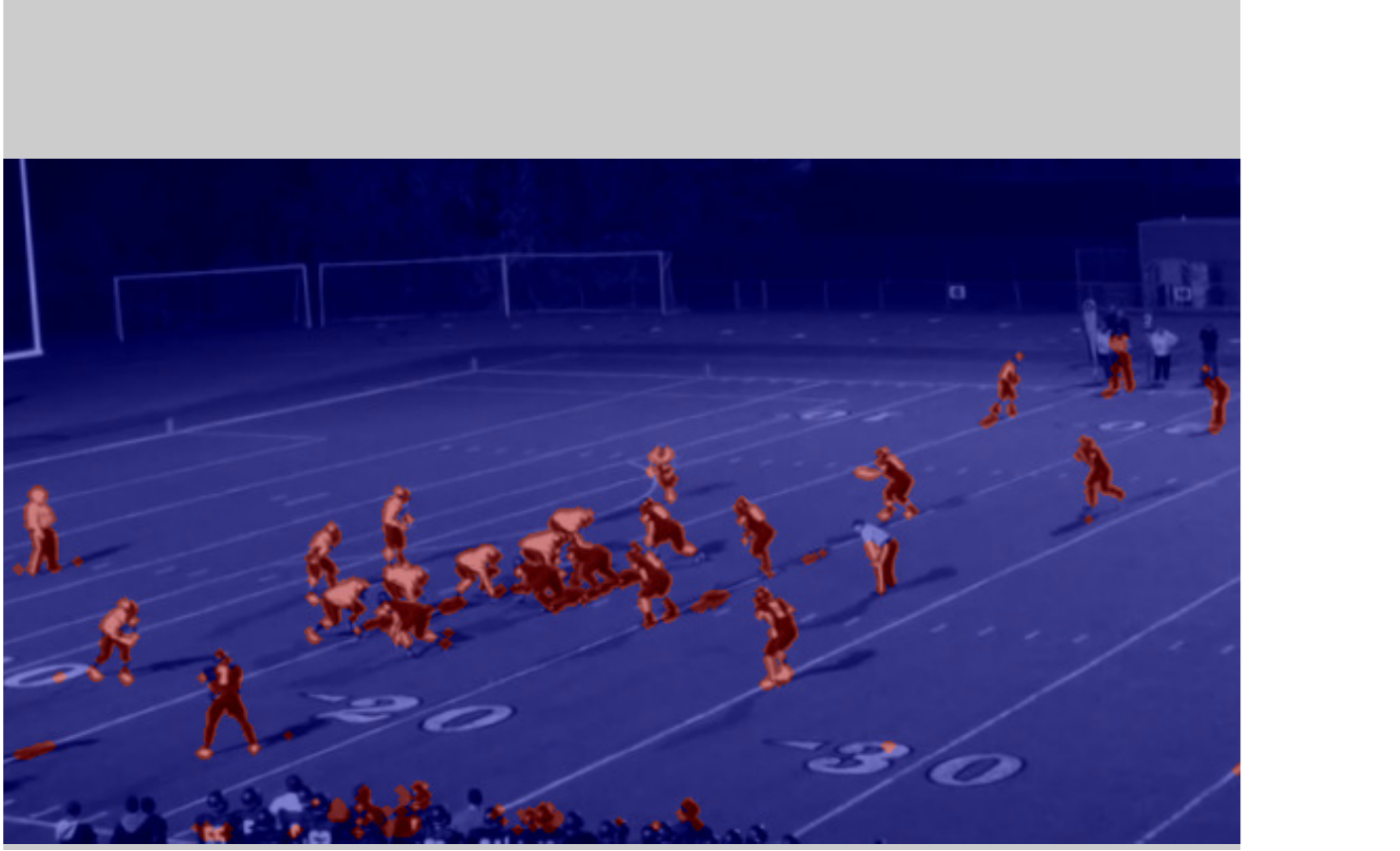} \\
%\ \ \ \ \ \ \ \ \ \ \ \ \ \ \ \ \ \ \ \small(c) & 
%\ \ \ \ \ \ \ \ \ \ \ \ \ \ \ \ \ \ \ \ \ \small(d)
\end{tabular}
%\caption{\small (a) Input frame, (b) reconstructed background, (c) difference of (a,b), (d) detected foreground via (c), overlaid on (a).}
\vspace{-4mm}
\caption{\small (a) Input frame, (b) reconstructed background, (c) difference of (a,b) on (a).}
\label{fig:fg}
\vspace{-7mm}
\end{figure}

\Paragraph{Foreground segmentation}
The reliable background reconstruction result $\mathbf{B}$ along with the GMC result of frame $\mathbf{I}^{(i)}$, e.g., $\p_i$, can be easily used to segment the foreground by thresholding the difference, $| \mathbf{B} - \mathbf{I}^{(i)}(\p_i) |$ (Fig.~\ref{fig:fg}).

%\vspace{3mm}
\Paragraph{Human action recognition}
State of the art human action recognition heavily relies on analysis of human motion. 
GMC helps to suppress camera motion and magnify human motion, making the motion analysis more feasible, which is clearly shown by the dense trajectories~\cite{dense} in Fig.~\ref{fig:densetraj}.
%For instance, the dense trajectories algorithm~\cite{dense} for motion analysis reveals its power when camera motion is compensated in the input video, either as pre-processing step, or internally~\cite{improveddense}.
%Otherwise, camera motion interferes with human motion, making the analysis problem very challenging.
%In~\cite{improveddense}, camera motion is compensated by detecting human and removing motion vectors due to human motion, and relying on RANSAC algorithm for outlier rejection. 
%However, this internal GMC requires accurate human detection, which has a high failure rate in videos in the wilds, specially for highly articulated human body in sports videos, and loses performance when number of false matches increases.
%Fig.~\ref{fig:densetraj} illustrates the difference of dense trajectories calculated on an input video with and without application of TRGMC.

\iffalse
\begin{figure} [t!]
\centering
%\begin{tabular}{p{3.8cm}p{3.5cm}}
\begin{tabular}{cc}
\includegraphics[trim=1.2cm 1.2cm 0.5cm 0cm,clip,height=1.7cm]{images/without.png} &
\includegraphics[trim=2.5cm 1.2cm 2cm 1.5cm,clip,height=1.7cm]{images/with.png} %\\
\end{tabular}
\caption{\small Dense trajectories of an original video (left), and the video processed by TRGMC (right).} %(For better visibility, boundary of frames are cropped.)}
\label{fig:densetraj}
%\figvspace
\vspace{-4mm}
\end{figure}
\fi

\begin{figure} [t!]
\centering
%\begin{tabular}{llll}
\begin{tabular}{p{0.5cm}p{3.4cm}p{0.5cm}p{3.4cm}}
\small (a) &
\includegraphics[trim=1.2cm 1.2cm 0.5cm 0cm,clip,height=1.6cm]{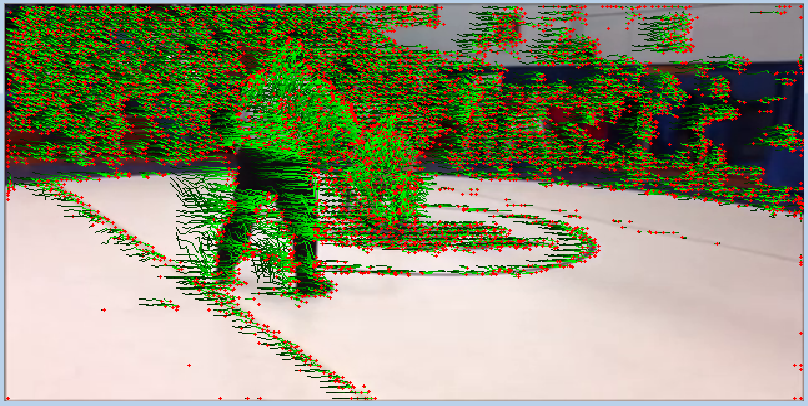} &
\small (b) &
\includegraphics[trim=2.5cm 1.2cm 2cm 1.5cm,clip,height=1.6cm]{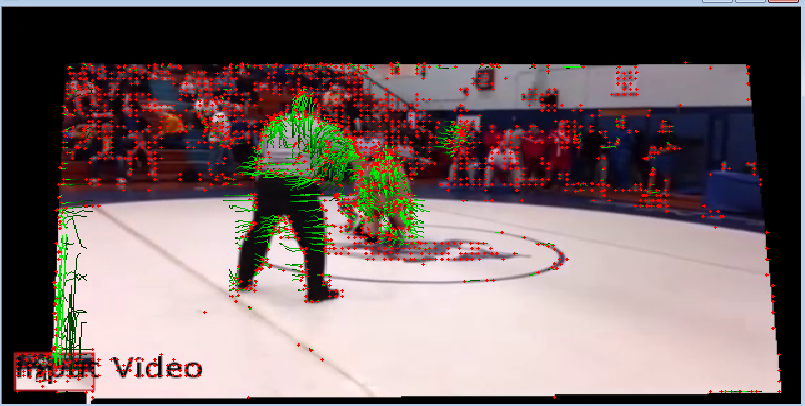} %\\
\end{tabular}
\vspace{-3mm}
\caption{\small Dense trajectories of the (a) original video, and (b) TRGMC-processed video.} %(For better visibility, boundary of frames are cropped.)}
\label{fig:densetraj}
\vspace{-1mm}
%\figvspace
\end{figure}

\begin{figure} [t!]
\vspace{-2mm}
\begin{center}
\centering
\begin{tabular}{@{}p{2.9cm}p{8cm}@{}}
\includegraphics[trim=0cm 0cm 0cm 0cm,clip,height=2.5cm]{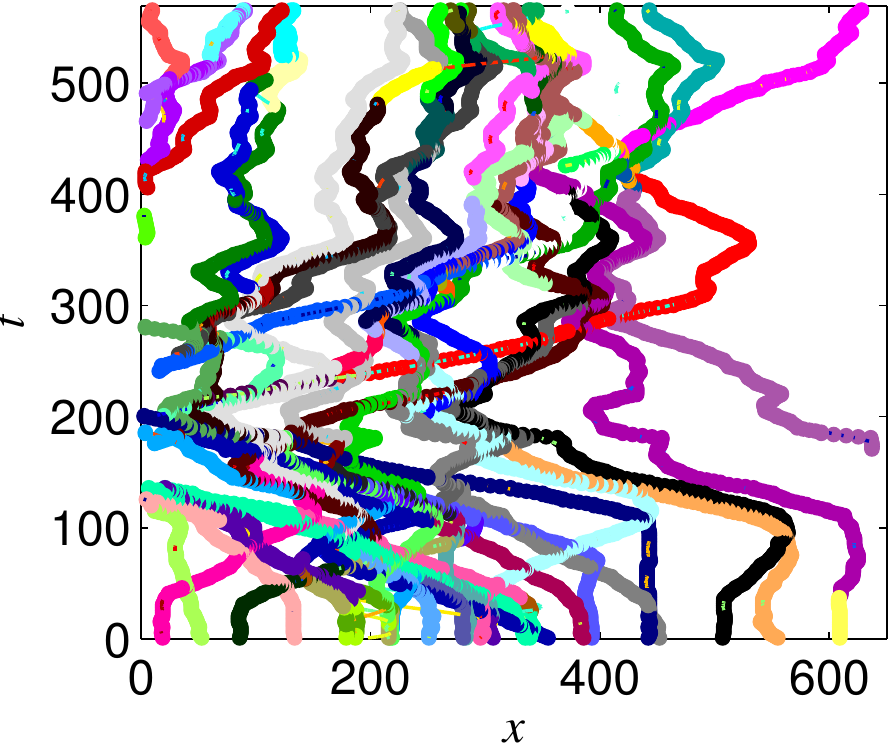} &
\includegraphics[trim=0cm 0cm 0cm 0cm,clip,height=2.5cm]{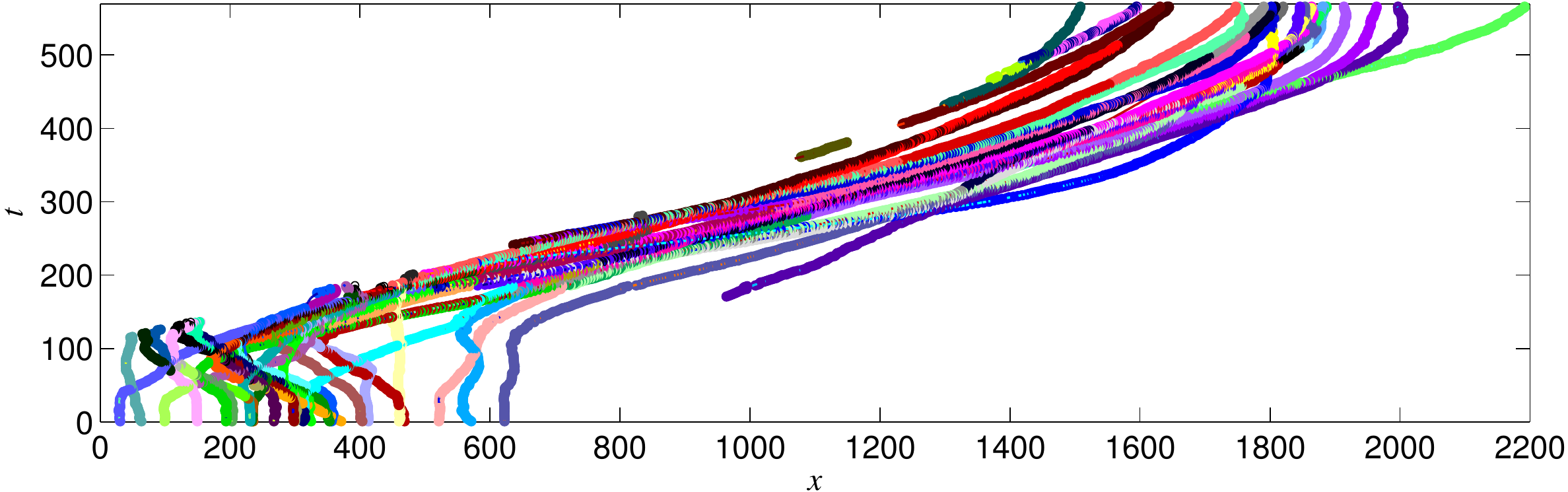} %\\
%\ \ \ \ \ \ \ \ \ \ \ \ \ \ \ \ \ \ \ \small(a) & 
%\ \ \ \ \ \ \ \ \ \ \ \ \ \ \ \ \ \ \ \ \ \small(b)
\end{tabular}
\end{center}
\vspace{-8mm}
\caption{\small Multi-player tracking using~\cite{andriyenko2012discrete} for a football video with camera panning to the right,  before (left) and after processing by TRGMC (right).} %Black color shows unassigned players.} 
\label{fig:tracking}
%\figvspace
\vspace{-5mm}
\end{figure}

\Paragraph{Multi-object Tracking (MOT)}
When appearance cues for tracking are ambiguous, e.g., tracking players in team sports like football, motion cues gain extra significance~\cite{lezama2011track,dicle2013way}. 
MOT is comprised of two tasks, data association by assigning each detection a label, and trajectory estimation -- both highly affected by camera motion. 
%However, motion of objects may be heavily impacted by dominant motion of camera.
TRGMC can be applied to remove camera motion and thus, revive the power of tracking algorithms relying on motion cues.
To verify the impact of TRGMC, we manually label the locations of all players in $566$ frames of a football video (the one in Fig.~\ref{fig:fg}) and use this ground truth detection results to study how MOT using~\cite{andriyenko2012discrete} benefits from TRGMC.
Fig.~\ref{fig:tracking} compares the trajectories of players over time with and without applying TRGMC. 
%In this figures, black color shows unassigned players.
Comparing number of label switches, this qualitatively demonstrates improvement of a challenging MOT scenario using TRGMC. 
%We report the Multi-Object Tracking Accuracy (MOTA)~\cite{bernardin2008evaluating} for both videos, which combines three metrics: missed targets, false alarms and identity switches, and then are normalized to a single value in the range $0-100\%$. 
%MOTA for the original is $63.79\%$, whereas the video pre-processed by TRGMC was able to outperform the former, with MOTA of $84.23\%$.
Also, the Multi-Object Tracking Accuracy (MOTA)~\cite{bernardin2008evaluating} achieved for the original video and the video processed by TRGMC are $63.79\%$ and $84.23\%$, respectively.

%------------------------------------------------------------%
\Section{Conclusions and Discussions}
We proposed a temporally robust global motion compensation (TRGMC) algorithm by joint alignment (congealing) of frames, in contrast to the common sequential scheme.
%This is done by dense connection of keypoints throughout all the frames and iteratively applying transformation on each frame such that the keypoints are spatially aligned.  
Despite complicated camera motions, TRGMC can remove the \textit{intentional} camera motion, such as pan, as well as \textit{unwanted} motion due to vibration on handheld cameras.
Experiments demonstrate that TRGMC outperforms existing GMC methods, and applications of TRGMC. 
%Beyond the motion panorama created by TRGMC, 
%We also illustrate various applications that might benefit from a robust and accurate GMC. %, e.g., human action recognition, background reconstruction, foreground detection, and object tracking.

The enabling assumption of TRGMC is that the camera motion in the direction of the optical axis is negligible. %for successful application 
For instance, TRGMC will not work properly on a video from a wearable camera of a pedestrian, since in the global coordinate the upcoming frames grow in size and cause computational and rendering problems. 
Similar to panorama images, the best results are achieved if the optical center of the camera has negligible movement during the capturing, making a homography-based approximation of camera motion appropriate. 
However, if the optical center moves in the perpendicular direction to the optical axis (e.g., a camera following a swimmer), TRGMC still works well, but rendering the results in the form of motion panorama will be degraded by parallax effect.

%$\Delta x_i(\p_i)$
%$\Delta y_i(\p_i)$
%$t$

%$\mathcal{W}_x(x_k^{(i)}, y_k^{(i)};  \p+\mathbf{\Delta p})$

%$\mathcal{W}_x(x_k^{(i)}, y_k^{(i)};  \p)$

%$\Delta x_i(\p+\mathbf{\Delta p})$

%$\Delta x_i(\p)$

\clearpage

\bibliographystyle{splncs}
\bibliography{egbib4,abbrev}

\begin{thebibliography}{10}

\bibitem{he2001fast}
He, Y., Feng, B., Yang, S., Zhong, Y.:
\newblock Fast global motion estimation for global motion compensation coding.
\newblock In: Proc. IEEE Int. Symp. Circuits and Systems (ISCAS). Volume~2.,
  IEEE (2001)  233--236

\bibitem{smolic2003long}
Smoli{\'c}, A., Vatis, Y., Schwarz, H., Wiegand, T.:
\newblock Long-term global motion compensation for advanced video coding.
\newblock In: ITG-Fachtagung Dortmunder Fernsehseminar. (2003)  213--216

\bibitem{bartoli2002video}
Bartoli, A., Dalal, N., Bose, B., Horaud, R.:
\newblock From video sequences to motion panoramas.
\newblock In: Proc. Conf. Motion and Video Computing Workshops, IEEE (2002)
  201--207

\bibitem{dense}
Wang, H., Klaser, A., Schmid, C., Liu, C.L.:
\newblock Action recognition by dense trajectories.
\newblock In: Proc. IEEE Conf. Computer Vision and Pattern Recognition (CVPR),
  IEEE (2011)  3169--3176

\bibitem{improveddense}
Wang, H., Schmid, C.:
\newblock Action recognition with improved trajectories.
\newblock In: Proc. Int. Conf. Computer Vision (ICCV), IEEE (2013)  3551--3558

\bibitem{monari2011real}
Monari, E., Pollok, T.:
\newblock A real-time image-to-panorama registration approach for background
  subtraction using pan-tilt-cameras.
\newblock In: Proc. IEEE Conf. Advanced Video and Signal Based Surveillance
  (AVSS), IEEE (2011)  237--242

\bibitem{sun2006better}
Sun, Y., Li, B., Yuan, B., Miao, Z., Wan, C.:
\newblock Better foreground segmentation for static cameras via new energy form
  and dynamic graph-cut.
\newblock In: Proc. Int. Conf. Pattern Recognition (ICPR). Volume~4., IEEE
  (2006)  49--52

\bibitem{wan2008new}
Wan, C., Yuan, B., Miao, Z.:
\newblock A new algorithm for static camera foreground segmentation via active
  coutours and {GMM}.
\newblock In: Proc. Int. Conf. Pattern Recognition (ICPR), IEEE (2008)  1--4

\bibitem{solera2015learning}
Solera, F., Calderara, S., Cucchiara, R.:
\newblock Learning to divide and conquer for online multi-target tracking.
\newblock arXiv preprint arXiv:1509.03956 (2015)

\bibitem{rgmc}
Safdarnejad, S.M., Liu, X., Udpa, L.:
\newblock Robust global motion compensation in presence of predominant
  foreground.
\newblock In: Proc. British Mach. Vision Conf. (BMVC). (2015)

\bibitem{bartoli2004motion}
Bartoli, A., Dalal, N., Horaud, R.:
\newblock Motion panoramas.
\newblock Computer Animation and Virtual Worlds \textbf{15}(5) (2004)  501--517

\bibitem{deniz2011fast}
D{\'e}niz, O., Bueno, G., Bermejo, E., Sukthankar, R.:
\newblock Fast and accurate global motion compensation.
\newblock Pattern Recognition \textbf{44}(12) (2011)  2887--2901

\bibitem{ransac}
Fischler, M.A., Bolles, R.C.:
\newblock Random sample consensus: a paradigm for model fitting with
  applications to image analysis and automated cartography.
\newblock ACM Commun. \textbf{24}(6) (1981)  381--395

\bibitem{chum2003locally}
Chum, O., Matas, J., Kittler, J.:
\newblock Locally optimized {RANSAC}.
\newblock Pattern Recognition (2003)  236--243

\bibitem{mle}
Torr, P.H., Zisserman, A.:
\newblock {MLESAC}: A new robust estimator with application to estimating image
  geometry.
\newblock Computer Vision and Image Understanding \textbf{78}(1) (2000)
  138--156

\bibitem{tordoff2005guided}
Tordoff, B.J., Murray, D.W.:
\newblock Guided-{MLESAC}: Faster image transform estimation by using matching
  priors.
\newblock IEEE Trans. Pattern Anal. Mach. Intell. \textbf{27}(10) (2005)
  1523--1535

\bibitem{ma2014robust}
Ma, J., Zhao, J., Tian, J., Yuille, A.L., Tu, Z.:
\newblock Robust point matching via vector field consensus.
\newblock IEEE Trans. Image Processing \textbf{23}(4) (2014)  1706--1721

\bibitem{li2010rejecting}
Li, X., Hu, Z.:
\newblock Rejecting mismatches by correspondence function.
\newblock Int. J. Computer Vision \textbf{89}(1) (2010)  1--17

\bibitem{heask}
Yan, Q., Xu, Y., Yang, X., Nguyen, T.:
\newblock {HEASK}: Robust homography estimation based on appearance similarity
  and keypoint correspondences.
\newblock Pattern Recognition \textbf{47}(1) (2014)  368--387

\bibitem{szpak2015robust}
Szpak, Z.L., Chojnacki, W., van~den Hengel, A.:
\newblock Robust multiple homography estimation: An ill-solved problem.
\newblock In: Proc. IEEE Conf. Computer Vision and Pattern Recognition (CVPR),
  IEEE (2015)  2132--2141

\bibitem{zuliani2005multiransac}
Zuliani, M., Kenney, C.S., Manjunath, B.:
\newblock The multi{RANSAC} algorithm and its application to detect planar
  homographies.
\newblock In: Proc. Int. Conf. Image Processing (ICIP). Volume~3., IEEE (2005)
  III--153

\bibitem{toldo2008robust}
Toldo, R., Fusiello, A.:
\newblock Robust multiple structures estimation with j-linkage.
\newblock In: Proc. European Conf. Computer Vision (ECCV).
\newblock Springer (2008)  537--547

\bibitem{ma2014mixture}
Ma, J., Chen, J., Ming, D., Tian, J.:
\newblock A mixture model for robust point matching under multi-layer motion.
\newblock PloS one \textbf{9}(3) (2014)  e92282

\bibitem{uemura2008feature}
Uemura, H., Ishikawa, S., Mikolajczyk, K.:
\newblock Feature tracking and motion compensation for action recognition.
\newblock In: Proc. British Mach. Vision Conf. (BMVC). (2008)  1--10

\bibitem{gao2011constructing}
Gao, J., Kim, S.J., Brown, M.S.:
\newblock Constructing image panoramas using dual-homography warping.
\newblock In: Proc. IEEE Conf. Computer Vision and Pattern Recognition (CVPR),
  IEEE (2011)  49--56

\bibitem{lin2011smoothly}
Lin, W.Y., Liu, S., Matsushita, Y., Ng, T.T., Cheong, L.F.:
\newblock Smoothly varying affine stitching.
\newblock In: Proc. IEEE Conf. Computer Vision and Pattern Recognition (CVPR),
  IEEE (2011)  345--352

\bibitem{zaragoza2014projective}
Zaragoza, J., Chin, T.J., Tran, Q.H., Brown, M.S., Suter, D.:
\newblock As-projective-as-possible image stitching with moving dlt.
\newblock IEEE Trans. Pattern Anal. Mach. Intell. \textbf{36}(7) (2014)
  1285--1298

\bibitem{lin2015adaptive}
Lin, C.C., Pankanti, S.U., Ramamurthy, K.N., Aravkin, A.Y.:
\newblock Adaptive as-natural-as-possible image stitching.
\newblock In: Proc. IEEE Conf. Computer Vision and Pattern Recognition (CVPR),
  IEEE (2015)  1155--1163

\bibitem{li2010generating}
Li, Y., Kang, S.B., Joshi, N., Seitz, S.M., Huttenlocher, D.P.:
\newblock Generating sharp panoramas from motion-blurred videos.
\newblock In: Proc. IEEE Conf. Computer Vision and Pattern Recognition (CVPR),
  IEEE (2010)  2424--2431

\bibitem{sawhney1998robust}
Sawhney, H.S., Hsu, S., Kumar, R.:
\newblock Robust video mosaicing through topology inference and local to global
  alignment.
\newblock In: Computer Vision—ECCV’98.
\newblock Springer (1998)  103--119

\bibitem{shum1998construction}
Shum, H.Y., Szeliski, R.:
\newblock Construction and refinement of panoramic mosaics with global and
  local alignment.
\newblock In: Computer Vision, 1998. Sixth International Conference on, IEEE
  (1998)  953--956

\bibitem{sakamoto2006homography}
Sakamoto, M., Sugaya, Y., Kanatani, K.:
\newblock Homography optimization for consistent circular panorama generation.
\newblock In: Advances in Image and Video Technology.
\newblock Springer (2006)  1195--1205

\bibitem{triggs1999bundle}
Triggs, B., McLauchlan, P.F., Hartley, R.I., Fitzgibbon, A.W.:
\newblock Bundle adjustment—a modern synthesis.
\newblock In: Vision algorithms: theory and practice.
\newblock Springer (1999)  298--372

\bibitem{el2011improved}
El-Saban, M., Izz, M., Kaheel, A., Refaat, M.:
\newblock Improved optimal seam selection blending for fast video stitching of
  videos captured from freely moving devices.
\newblock In: Proc. Int. Conf. Image Processing (ICIP), IEEE (2011)  1481--1484

\bibitem{perazzi2015panoramic}
Perazzi, F., Sorkine-Hornung, A., Zimmer, H., Kaufmann, P., Wang, O., Watson,
  S., Gross, M.:
\newblock Panoramic video from unstructured camera arrays.
\newblock In: Computer Graphics Forum. Volume~34., Wiley Online Library (2015)
  57--68

\bibitem{zeng2009depth}
Zeng, W., Zhang, H.:
\newblock Depth adaptive video stitching.
\newblock In: Proc. IEEE Conf. Computer and Information Science (ICIS), IEEE
  (2009)  1100--1105

\bibitem{jiang2015video}
Jiang, W., Gu, J.:
\newblock Video stitching with spatial-temporal content-preserving warping.
\newblock In: Proc. IEEE Conf. Computer Vision and Pattern Recognition
  Workshops (CVPRW), IEEE (2015)  42--48

\bibitem{ibrahim2012automatic}
Ibrahim, M.T., Hafiz, R., Khan, M.M., Cho, Y., Cha, J.:
\newblock Automatic reference selection for parametric color correction schemes
  for panoramic video stitching.
\newblock In: Advances in Visual Computing.
\newblock Springer (2012)  492--501

\bibitem{gleicher2008re}
Gleicher, M.L., Liu, F.:
\newblock Re-cinematography: Improving the camerawork of casual video.
\newblock ACM Transactions on Multimedia Computing, Communications, and
  Applications (TOMM) \textbf{5}(1) (2008) ~2

\bibitem{learned2006data}
Learned-Miller, E.G.:
\newblock Data driven image models through continuous joint alignment.
\newblock IEEE Trans. Pattern Anal. Mach. Intell. \textbf{28}(2) (2006)
  236--250

\bibitem{Liu2009b}
Liu, X., Tong, Y., Wheeler, F.W.:
\newblock Simultaneous alignment and clustering for an image ensemble.
\newblock In: Proc. Int. Conf. Computer Vision (ICCV), IEEE (2009)  1327--1334

\bibitem{cox2008least}
Cox, M., Sridharan, S., Lucey, S., Cohn, J.:
\newblock Least squares congealing for unsupervised alignment of images.
\newblock In: Proc. IEEE Conf. Computer Vision and Pattern Recognition (CVPR),
  IEEE (2008)  1--8

\bibitem{huang2012learning}
Huang, G., Mattar, M., Lee, H., Learned-Miller, E.G.:
\newblock Learning to align from scratch.
\newblock In: Advances in Neural Information Processing Systems (NIPS). (2012)
  764--772

\bibitem{lankinen2011local}
Lankinen, J., K{\"a}m{\"a}r{\"a}inen, J.K.:
\newblock Local feature based unsupervised alignment of object class images.
\newblock In: Proc. British Mach. Vision Conf. (BMVC). Volume~1. (2011) ~5

\bibitem{lucey2013fourier}
Lucey, S., Navarathna, R., Ashraf, A.B., Sridharan, S.:
\newblock Fourier lucas-kanade algorithm.
\newblock IEEE Trans. Pattern Anal. Mach. Intell. \textbf{35}(6) (2013)
  1383--1396

\bibitem{cox2009least}
Cox, M., Sridharan, S., Lucey, S., Cohn, J.:
\newblock Least-squares congealing for large numbers of images.
\newblock In: Proc. Int. Conf. Computer Vision (ICCV), IEEE (2009)  1949--1956

\bibitem{shokrollahi2015unsupervised}
Shokrollahi~Yancheshmeh, F., Chen, K., Kamarainen, J.K.:
\newblock Unsupervised visual alignment with similarity graphs.
\newblock In: Proc. IEEE Conf. Computer Vision and Pattern Recognition (CVPR),
  IEEE (2015)  2901--2908

\bibitem{surf}
Bay, H., Tuytelaars, T., Van~Gool, L.:
\newblock {SURF}: Speeded up robust features.
\newblock In: Proc. European Conf. Computer Vision (ECCV).
\newblock Springer (2006)  404--417

\bibitem{lowe2004distinctive}
Lowe, D.G.:
\newblock Distinctive image features from scale-invariant keypoints.
\newblock Int. J. Computer Vision \textbf{60}(2) (2004)  91--110

\bibitem{svw}
Safdarnejad, S.M., Liu, X., Udpa, L., Andrus, B., Wood, J., Craven, D.:
\newblock Sports videos in the wild ({SVW}): A video dataset for sports
  analysis.
\newblock In: Proc. Int. Conf. Automatic Face and Gesture Recognition (FG),
  IEEE (2015)  1--7

\bibitem{ucf101}
Soomro, K., Zamir, A.R., Shah, M.:
\newblock {UCF101}: A dataset of 101 human actions classes from videos in the
  wild.
\newblock arXiv preprint arXiv:1212.0402 (2012)

\bibitem{steedly2005efficiently}
Steedly, D., Pal, C., Szeliski, R.:
\newblock Efficiently registering video into panoramic mosaics.
\newblock In: Proc. Int. Conf. Computer Vision (ICCV). Volume~2., IEEE (2005)
  1300--1307

\bibitem{lezama2011track}
Lezama, J., Alahari, K., Sivic, J., Laptev, I.:
\newblock Track to the future: Spatio-temporal video segmentation with
  long-range motion cues.
\newblock In: Proc. IEEE Conf. Computer Vision and Pattern Recognition (CVPR),
  IEEE (2011)

\bibitem{dicle2013way}
Dicle, C., Camps, O., Sznaier, M.:
\newblock The way they move: tracking multiple targets with similar appearance.
\newblock In: Proc. Int. Conf. Computer Vision (ICCV), IEEE (2013)  2304--2311

\bibitem{andriyenko2012discrete}
Andriyenko, A., Schindler, K., Roth, S.:
\newblock Discrete-continuous optimization for multi-target tracking.
\newblock In: Proc. IEEE Conf. Computer Vision and Pattern Recognition (CVPR),
  IEEE (2012)  1926--1933

\bibitem{bernardin2008evaluating}
Bernardin, K., Stiefelhagen, R.:
\newblock Evaluating multiple object tracking performance: the {CLEAR MOT}
  metrics.
\newblock J. Image and Video Process. \textbf{2008} (2008) ~1

\end{thebibliography}
\end{document}